\pdfoutput=1
\PassOptionsToPackage{table,dvipsnames}{xcolor}
\documentclass[11pt, a4paper, logo, copyright]{gdm_format}

\usepackage[authoryear, sort&compress, round]{natbib}
\usepackage{tabularx}
\usepackage{threeparttable}
\usepackage{array,multirow,nicefrac,placeins}
\usepackage[linesnumbered,noend]{algorithm2e}
\usepackage[caption=false,font=footnotesize]{subfig}
\usepackage[most]{tcolorbox}
\makeatletter
\def\HyCitePfx{}
\def\hyper@natlinkstart#1{\Hy@backout{#1}\hyper@linkstart{cite}{cite.\HyCitePfx#1}\def\hyper@nat@current{#1}}
\def\hyper@natlinkbreak#1#2{\hyper@linkend#1\hyper@linkstart{cite}{cite.\HyCitePfx#2}}
\def\hyper@natanchorstart#1{\Hy@raisedlink{\hyper@anchorstart{cite.\HyCitePfx#1}}}
\makeatother
\let\cite\citep
\usepackage{bibunits}
\defaultbibliographystyle{iclr2021_conference}
\defaultbibliography{refs}
\usepackage{wrapfig}
\usepackage{fontawesome5}
\definecolor{citeblue}{RGB}{16,52,130}
\hypersetup{allcolors=citeblue}
\newcommand{\methodname}{OpenArchEvo}

\renewcommand{\copyrightext}{\footerfont This work has been submitted to the IEEE for possible publication. Copyright may be transferred without notice, after which this version may no longer be accessible.}

\title{Large Language Model-Guided Evolutionary Discovery of Native Neural Architectures for Spiking Sequence Modeling}

\author[]{Ruoyu Zhao\textsuperscript{*}}
\author[]{Jiaqi Wu\textsuperscript{*}}
\author[]{Chenyu Zhu}
\author[]{Zhichao Lu}
\affil[]{Department of Computer Science, City University of Hong Kong}
\affil[]{\texttt{\{ruoyuzhao8-c, jwu395-c, chenyuzhu9-c\}@my.cityu.edu.hk}, \texttt{zhichao.lu@cityu.edu.hk}}

\correspondingauthor{Zhichao Lu (zhichao.lu@cityu.edu.hk). \textsuperscript{*}Equal contribution.}

\begin{document}

\begin{abstract}
Spiking neural networks (SNNs) offer low-energy sequence modeling through sparse, event-driven computation. However, interactions among spike encoding, neuronal dynamics, and information propagation complicate architecture design.
Existing SNN sequence models often adapt artificial neural network (ANN) architectures designed for real-valued activations, potentially underusing spike-based communication and temporal state updates, motivating automated discovery of native SNN architectures.
Most evolutionary neural architecture search (ENAS) methods operate within predefined configuration spaces, limiting discovery to mechanisms expressible within those spaces. We introduce \methodname, which uses large language models (LLMs) to evolve executable architecture code in an open program space under spiking-projection constraints.
In this space, code differences need not reflect architectural novelty, while direct performance evaluation requires costly training. We construct a three-view representation spanning code, design rationale, and a behavioral fingerprint to support novelty estimation and performance prediction. The search treats predicted performance and estimated novelty as two objectives, using surrogate predictions to select candidates for expensive training evaluations.
With an estimated candidate-training cost of 132 V100 GPU-days, the search uncovers multiple native SNN architectures, exemplified by three designs featuring mechanisms such as spike-activity-dependent control of state updates and residual pathways.
The discovered NeuroGate surpasses the ANN DeltaNet on WikiText-103, and the discovered architectures reduce estimated architecture-level arithmetic energy by up to $50.6\times$ (LoopMem) relative to a common dense Transformer (ANN) baseline.
\par\medskip
{\centering\urlstyle{same}%
\raisebox{-0.1em}{\faGithub}~Code: \url{https://github.com/Optima-CityU/OpenArchEvo}\par\smallskip
\raisebox{-0.15em}{\includegraphics[height=1em]{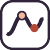}}~Project page: \url{https://optima-cityu.github.io/OpenArchEvo/}\par}
\medskip
\end{abstract}

\maketitle

\begin{bibunit}
\section{Introduction}
\label{sec:intro}

\begin{figure*}[t]
\centering
\includegraphics[width=0.94\linewidth]{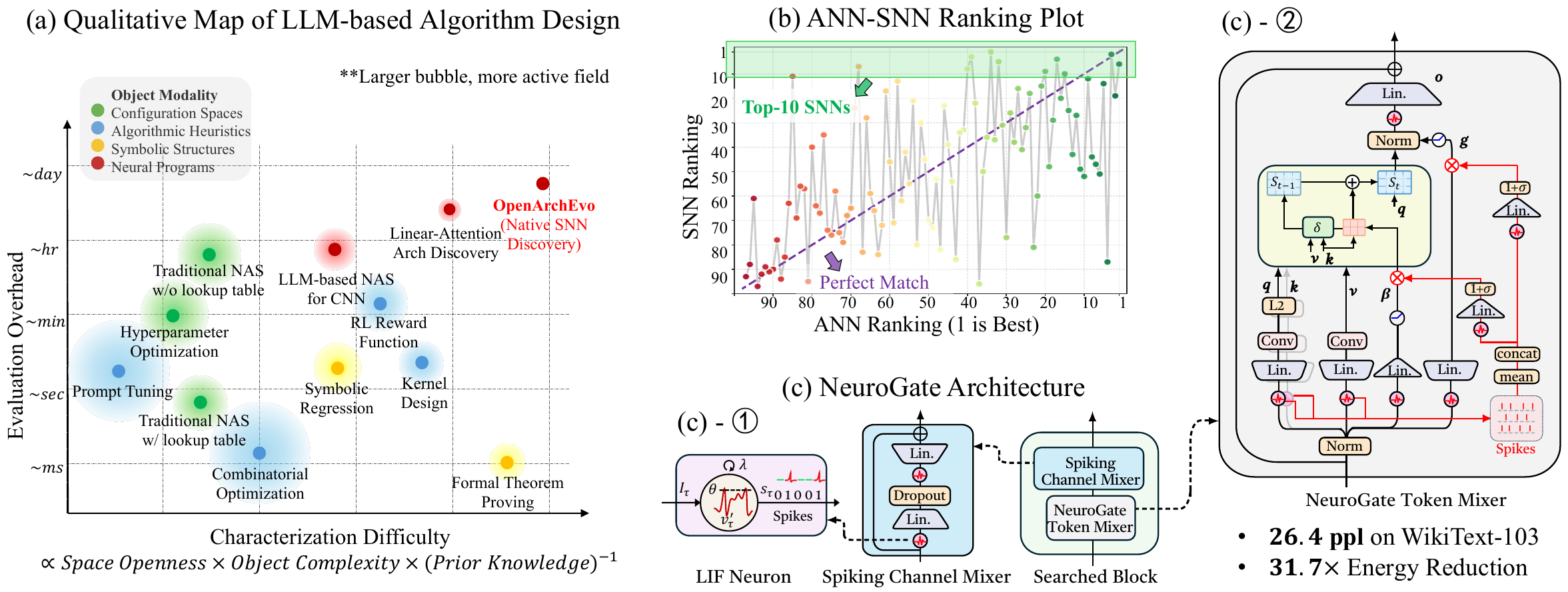}
\caption{Motivation and result preview for native SNN architecture discovery. \textbf{(a) A qualitative map} of LLM-based algorithm design by candidate-characterization difficulty and evaluation cost. \textbf{(b) Partial rank agreement} is shown by separately trained ANN and SNN implementations of the 97 ASI-Arch paired architectures on WikiText-2 (Kendall $\tau_b=0.2997$; Appendix~\ref{app:ann_snn_mismatch}). \textbf{(c) NeuroGate} achieves 26.4 WikiText-103 perplexity and a 31.7$\times$ estimated architecture-level arithmetic-energy reduction relative to a common dense Transformer (ANN) baseline (Table~\ref{tab:snn}), under spiking-projection constraints (Appendix~\ref{app:fully_snn_definition}). Red arrows mark the added pathway through which spike activity modulates the recurrent update and output gating. }
\label{fig:motivation}\vspace{-12pt}
\end{figure*}

Spiking neural networks (SNNs) communicate through sparse, event-driven spikes, offering a promising route to low-energy sequence modeling. Binary spike signals interact with continuous states that retain and integrate information over time. Architecture determines how spike activity updates these states and how stored information is propagated, shaping predictive performance and opportunities for sparse computation. Existing SNN sequence models include both adaptations of artificial neural network (ANN) architectures and mechanisms designed specifically for spiking computation~\cite{zhu2025spikegpt,zhong2024spike,shen2025spikingssms}. However, architectural choices effective in ANNs need not retain their advantages after adaptation to spiking computation. In our WikiText-2 comparison, ANN and SNN implementations of the 97 ASI-Arch paired architectures agree only partially in rank (Fig.~\ref{fig:motivation}(b)), motivating direct evaluation under spiking computation. The interactions among spike activity, state updates, and information pathways make useful combinations difficult to anticipate. This motivates an AutoML research question: \emph{how can search build on expert designs to discover native SNN architectures that match or exceed ANN performance while preserving substantial energy advantages?}

Neural architecture search (NAS), particularly evolutionary NAS (ENAS), offers a route to this automated exploration~\cite{liu2023enassurvey,li2024edlsurvey}. Most ENAS methods vary operators, connections, and hyperparameters within predefined configuration spaces. These spaces permit new combinations of known components, but mechanisms outside their allowed operations and composition rules require redesigning the search representation. Large language models (LLMs) can generate and modify executable architecture code, enabling changes to internal computations and module interactions. Existing LLM-based NAS already supports architecture discovery through code evolution, including quality-diversity search and novelty checks~\cite{chen2023evoprompting,nasir2024llmatic,liu2025alphago}. Greater freedom to generate candidates, however, does not by itself yield effective architectural discovery. Figure~\ref{fig:motivation}(a) situates this problem within LLM-based algorithm design: code differences need not reflect meaningful architectural differences, while directly measuring task performance requires costly candidate training. The challenge is to characterize generated SNN architectures for novelty estimation and performance prediction, guiding training toward promising candidates while retaining distinct architectural alternatives.

To address this challenge, we introduce \textbf{\methodname}, an LLM-guided evolutionary method for native SNN architecture discovery. It searches over an open code implementation space for SNN blocks within a fixed outer model structure, subject to interface, causality, and spiking-projection constraints. We estimate each candidate's architectural novelty relative to other architectures considered during search through three complementary views: code describes computational structure, the design rationale states the intended design, and a behavioral fingerprint combines architecture statistics with network responses measured before weight training. The fingerprint also supports surrogate performance prediction; predicted performance and estimated novelty form two objectives for selecting candidates for training. One discovered architecture, NeuroGate, features spike-activity-dependent modulation of recurrent updates and output gating. On WikiText-103, the discovered SNNs outperform the listed SNN baselines, and NeuroGate (Fig.~\ref{fig:motivation}(c)) attains 26.4 WikiText-103 perplexity versus 27.5 for ANN DeltaNet (lower is better). The discovered architectures reduce estimated architecture-level arithmetic energy by up to $50.6\times$ (LoopMem) relative to a common dense Transformer (ANN) baseline (Table~\ref{tab:snn}).

Our main contributions are as follows:
\begin{itemize}[leftmargin=*,topsep=2pt,itemsep=3pt,parsep=0pt]
\item \textbf{Native SNN architectures.} We discover SNN architectures with evolved elements native to spiking computation, including spike-activity-dependent control in NeuroGate and HomeoResSSM and a state-norm feedback in LoopMem that benefits the spiking implementation, and examine these elements through SNN ablations and a non-spiking comparison.

\item \textbf{Three-view architecture representation.} We construct a three-view representation using code, design rationale, and numerical behavioral fingerprints of architectural structure and behavior. We compare candidates using these views to estimate each candidate's architectural novelty, while reusing the fingerprints for performance prediction.

\item \textbf{LLM-guided evolutionary discovery.} We integrate this characterization into LLM-guided evolution over an open code implementation space, using surrogate-predicted performance and estimated novelty as two objectives to select architectures for training within a limited budget. Search diagnostics and ablations examine surrogate guidance and diversity maintenance.
\end{itemize}

\section{Related Work}
\label{sec:related}

We review SNN architecture design, evolutionary neural architecture search, and representations for candidate comparison and performance prediction.

\subsection{Spiking Sequence Models}
\label{subsec:snn}

Spiking neurons combine binary outputs with continuous membrane states that retain temporal information. Architectural choices determine how these signals are transformed and propagated. In vision models, Spikformer uses spike-based attention without softmax~\cite{zhou2023spikformer}, while Spike-driven Transformer places residual connections before spiking activations to preserve binary communication~\cite{yao2023spikedriven}. These designs show how adapting information pathways to spiking computation can support performance with less computation.

For sequence modeling, SpikeGPT combines recurrent processing with binary spiking activations for language generation~\cite{zhu2025spikegpt}. The SpikingSSMs architecture applies leaky integrate-and-fire (LIF) dynamics to state-space outputs, combining temporal memory with sparse synaptic computation~\cite{shen2025spikingssms}. Dyn-SSM incorporates refractory LIF neurons with soft reset, preserving residual membrane potential after firing~\cite{zhong2024spike}.

Automated methods also demonstrate the value of SNN-specific architectural exploration. SNASNet searches forward and temporal feedback connections, while MSE-NAS explores neuron operations and multiscale connectivity~\cite{kim2022snasnet,pan2025msenas}. AutoSNN considers accuracy and spike count and reports improvements over its handcrafted baselines~\cite{na2022autosnn}. For spiking language models, EQ-SpikeLM combines evolutionary channel pruning with subsequent post-training quantization~\cite{zhang2026eqspikelm}. \emph{Further couplings among spike activity, sequence-state updates, and residual pathways offer opportunities to extend this accumulated design knowledge.}

\subsection{Evolutionary Neural Architecture Search}
\label{subsec:enas}

Evolutionary neural architecture search (ENAS) uses population-based variation and selection to optimize neural architectures~\cite{liu2023enassurvey,li2024edlsurvey}. Most methods vary operators, connections, widths, and depths within predefined representations. For example, EvoCNN evolves variable-length sequences of convolutional, pooling, and fully connected layers~\cite{sun2020evocnn}, while NSGA-Net evolves CNN blocks through predefined operation and connection choices~\cite{lu2021nsganet}. These encodings support new combinations of existing building blocks, but a mechanism outside their permitted operations and composition rules requires redesigning the search representation.

Genetic programming and grammar-based NAS support variable computational structures assembled from reusable primitives. CGP-CNN evolves convolutional architectures from layer-level components, while einspace uses typed primitives to define a broader compositional space~\cite{suganuma2017cgpcnn,ericsson2024einspace}. AutoML-Zero further explores model computation and learning rules constructed from basic mathematical operations~\cite{real2020automlzero}. High-level primitives supply architectural priors through predefined computational forms; finer primitives expose more computational choices but can make useful designs harder to find. Even when composition obeys type and interface constraints, finding useful architectures can require many expensive evaluations. This motivates using architectural knowledge to guide the generation and modification of candidate architecture code.

LLM-guided evolutionary search has become a general approach to automated algorithm design~\cite{wu2025ecllm,ma2026metabbo}, evolving heuristics, metaheuristics, and scientific programs~\cite{romera2024mathematical,liu2024evolution,vanstein2025llamea,novikov2025alphaevolve}; ShinkaEvolve, for example, combines code-embedding similarity screening with an LLM novelty judge to reject redundant candidate programs~\cite{lange2025shinka}. For neural architectures, LLMs use pretrained knowledge and natural-language design goals, constraints, and principles to guide configuration search or generate architecture code. LLMENAS adapts fitness functions within a predefined cell search space~\cite{lai2026llmenas}, while Design Principle Transfer uses natural-language principles to narrow subsequent searches~\cite{zhou2025designtransfer}. EvoPrompting evolves complete classifiers and, for graph networks, selected computations within a fixed processor~\cite{chen2023evoprompting}. LLMatic combines network-code evolution with quality-diversity search for image classification~\cite{nasir2024llmatic}. ASI-Arch evolves attention-layer implementations under interface and computation constraints; it checks novelty through rationale retrieval and LLM judgments before training, and includes LLM-assessed architectural quality in its fitness~\cite{liu2025alphago}; Genesys discovers language-model architectures with LLM agents on a genetic-programming backbone~\cite{cheng2025genesys}. These code-based approaches establish architecture discovery with mechanisms for diversity and novelty. \emph{Our work combines population-relative novelty and surrogate-predicted performance to select SNN architecture programs for training; Section~\ref{sec:qd_background} details the representations supporting these decisions.}

\subsection{Architecture Representation for Evolutionary Search}
\label{sec:qd_background}

A search encoding specifies how architectures can be constructed; candidate descriptors summarize the resulting architectures for comparison and prediction. Implementation-level comparisons include the lexical, syntactic, and data-flow matches measured by CodeBLEU~\cite{ren2020codebleu}. Natural-language descriptions expose design intent, as in Evolution of Heuristics (EoH), which jointly evolves heuristic ideas and executable implementations~\cite{liu2024evolution}. Execution-based descriptions capture observed behavior: phenotypic characterization records program responses to probe cases and supports surrogate modeling~\cite{hildebrandt2015surrogates}, while BehaveSim compares intermediate solution trajectories~\cite{zhang2026rethinking}. These views provide complementary, partial evidence: similar code need not implement the same computation, stated design intent need not match the implementation, and finite probes reveal only part of a program's behavior.

For neural architectures, structural statistics and initialization-time forward and backward probes provide information before candidate training. Training-free proxies estimate trained performance from initialization-time activations or gradients without optimizing candidate weights~\cite{abdelfattah2021zerocost}. DCL-ENAS pretrains an architecture encoder without performance labels, then contrastively fine-tunes a predictor on evaluated architectures to predict their relative performance~\cite{zhang2026dclenas}.
In expressive grammar-based NAS, Transferrable Surrogates builds transferable predictors from training-free proxies and neural graph features, or from a fine-tuned language model, and uses them to filter candidates or serve directly as search objectives~\cite{qin2025transferrable}.
Accurate performance prediction alone does not validate distances between architecture representations as measures of novelty.

Descriptors also determine which differences evolutionary search rewards. Novelty search rewards distance from previously observed behaviors, while MAP-Elites retains high-performing solutions within descriptor-defined regions~\cite{lehman2011novelty,mouret2015illuminating}. In LLMatic, discussed in Section~\ref{subsec:enas}, the network archive uses width-to-depth ratio and floating-point operations (FLOPs) as descriptors~\cite{nasir2024llmatic}. BOP-Elites models both quality and descriptors with surrogates to select evaluations when these quantities are expensive to obtain~\cite{kent2025bopelites}. For SNNs, architectures with similar size and computational cost can nevertheless differ in how spike activity interacts with state updates and gating. \emph{We estimate novelty relative to a reference population by comparing code, design rationale, and a behavioral fingerprint combining architecture statistics with initialization-time probes. The same fingerprint supports performance prediction, and both estimates guide the selection of candidates for expensive training evaluations.}

\begin{figure*}[t]
\centering
\includegraphics[width=0.97\linewidth]{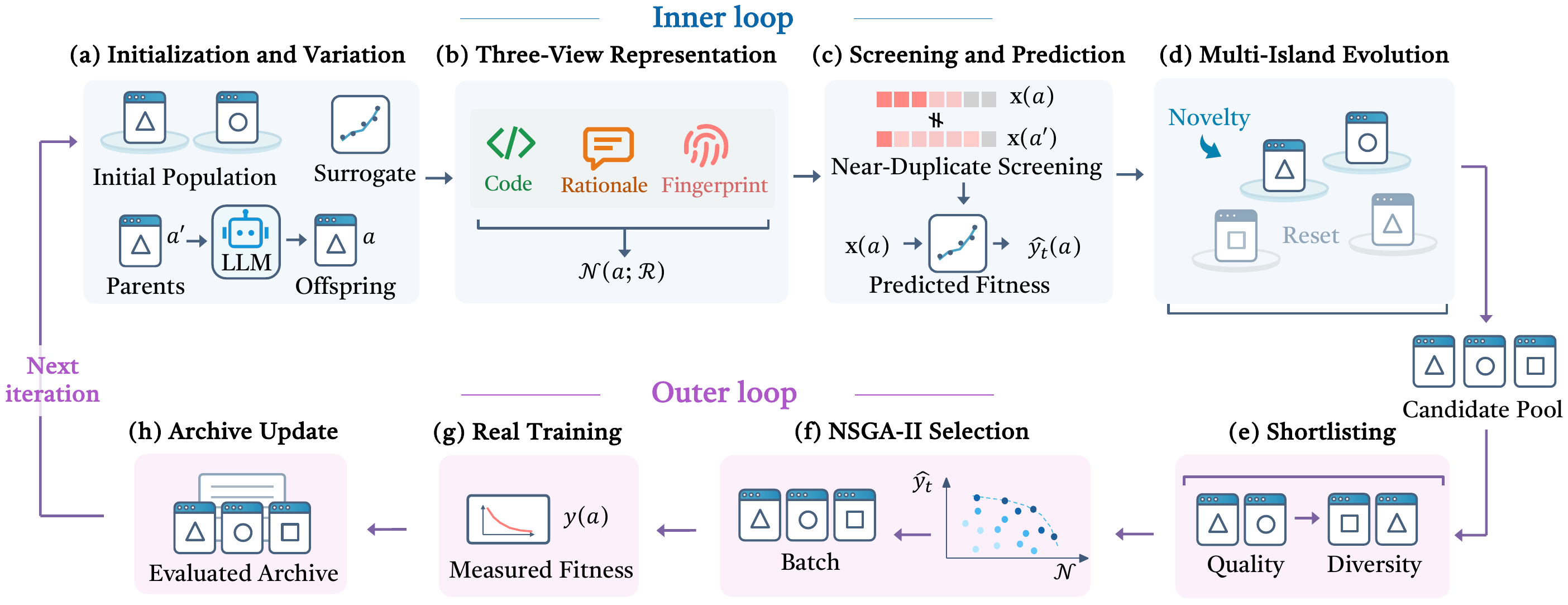}
\caption{Overview of \methodname. The inner loop generates feasible SNN architecture programs and uses surrogate predictions to guide evolution. The outer loop selects a small training batch using predicted performance and estimated novelty, then adds measured outcomes to the evaluated archive. Algorithms~\ref{alg:sandnas} and~\ref{alg:algobleu} in Appendix~\ref{app:algorithms} give the procedures.}
\label{fig:pipeline}
\end{figure*}

\section{Method}
\label{sec:method}

\methodname\ discovers SNN architecture programs by coupling program-level variation with three-view architecture representation and surrogate-assisted evolution (Fig.~\ref{fig:pipeline}). Code, design rationale, and a behavioral fingerprint provide complementary descriptions for comparing generated architectures. The fingerprint also supports near-duplicate screening and performance prediction, allowing an inner evolutionary loop to explore many candidates before an outer loop allocates expensive training evaluations. Performance--novelty selection encourages the retention of distinct architectural directions, and measured training outcomes guide subsequent search.

\subsection{Preliminaries on SNNs}
\label{sec:program-space}

An SNN processes information through interconnected spiking neurons, whose internal states evolve over time and whose outputs are discrete binary events~\cite{neftci2019surrogate}. In a discrete-time description, a neuron's binary output records whether it fires at each step, while its membrane potential remains continuous-valued. A common example is the leaky integrate-and-fire (LIF) neuron. In its hard-reset form, the pre-reset potential $\tilde v_\tau$, spike $s_\tau$, and membrane state $v_\tau$ at step $\tau$ satisfy
\begin{equation}
\tilde v_\tau=\lambda v_{\tau-1}+I_\tau,\quad
s_\tau=\mathbb I[\tilde v_\tau\geq\theta],\quad
v_\tau=(1-s_\tau)\tilde v_\tau,
\label{eq:lif}
\end{equation}
where \(I_\tau\), \(\lambda\), and \(\theta\) denote input current, leakage factor, and firing threshold, and \(\mathbb I[\cdot]\) denotes the indicator function. The neuron integrates the current with its decayed membrane state, emits a spike when the resulting potential reaches or exceeds the threshold, and resets its state to zero after firing.

Within the network, incoming spikes contribute to a neuron's input current through weighted synaptic connections. The resulting spike response depends on both the incoming signals and the neuron's preceding membrane state. Binary synaptic inputs also permit weighted sums to be evaluated by accumulating the weights associated with active spikes, providing opportunities for sparse computation~\cite{davies2018loihi,horowitz2014computing}.

\subsection{Problem Formulation}
\label{sec:formulation}

Let $\mathcal A$ denote the search space of admissible SNN block architectures, each represented by executable code and feeding binary spikes to its parameter-dominant feature projections (the spiking-projection constraint; Appendix~\ref{app:fully_snn_definition}). Architecture performance is task-dependent, so our main search combines WikiText-2 language modeling and ListOps hierarchical expression evaluation~\cite{merity2017pointer,nangia-bowman-2018-listops} to assess each design beyond a single task. Their weak rank agreement in our cross-task comparison supports their use as complementary evaluation signals (Appendix~\ref{app:cap_ab}). For $a\in\mathcal A$, let $w$ collect the separate trainable parameter vectors of the full models instantiated with $a$ for the evaluation tasks, and let $\mathcal L_{\mathrm{train}}(a,w)$ sum their training losses. We aggregate their evaluation scores into a single performance objective $y(a,w)$, with larger values indicating better performance. Let $\mathcal R$ be a finite set of reference architectures, held fixed for each ranking, and let $\mathcal N(a;\mathcal R)$ denote the estimated architectural novelty of $a$ relative to $\mathcal R$. To encourage distinct architectural alternatives alongside performance, we model discovery as a multiobjective optimization problem following NAS formulations~\cite{lu2021nsganet,lu2023evoxbench} and novelty-based multiobjectivization~\cite{mouret2012encouraging}, mathematically as follows:
\begin{equation}
\begin{aligned}
\underset{a\in\mathcal A}{\operatorname{maximize}}\quad &
\bigl(y(a,w^*(a)),\mathcal N(a;\mathcal R)\bigr),\\
\text{subject to}\quad &
w^*(a)\in\operatorname*{arg\,min}_{w}\mathcal L_{\mathrm{train}}(a,w).
\end{aligned}
\label{eq:search-objectives}
\end{equation}

In practice, fixed, finite training protocols approximate the lower-level optimization. Measuring a candidate's fitness in the main search requires separate training on both datasets, increasing evaluation cost relative to training the same candidate on either task alone. Novelty can instead be estimated before weight training. To limit evaluation cost, each search run evaluates at most $B$ additional architectures under its prescribed task and training protocol, excluding those evaluated initially.

\begin{figure*}[t]
\centering
\includegraphics[width=\linewidth]{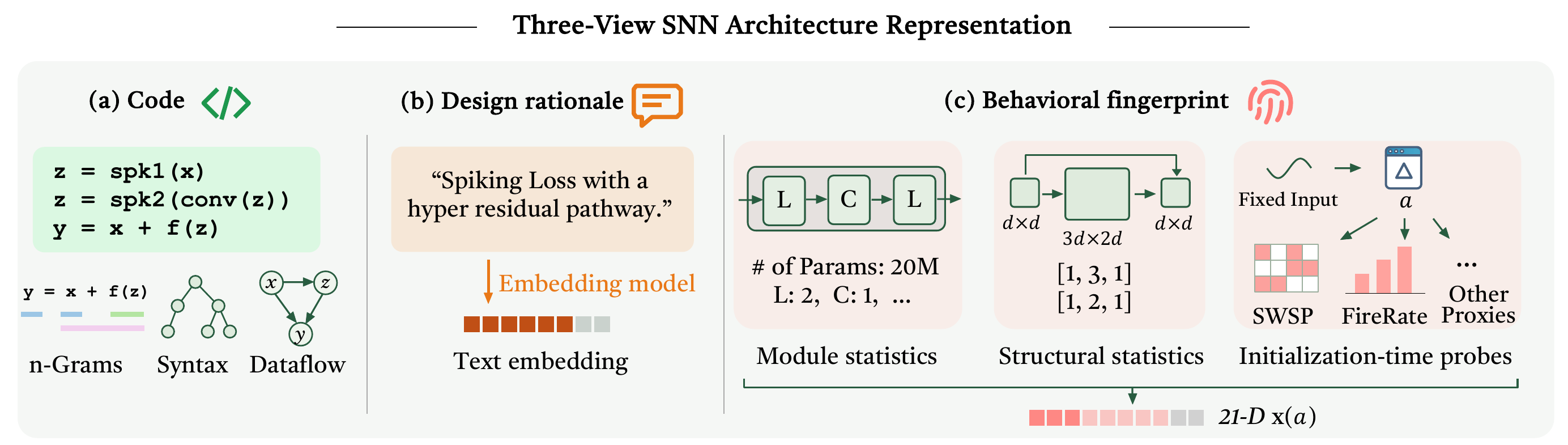}
\caption{Three-view representation of an SNN architecture program. Code and design rationale describe its implementation and intended choices; a behavioral fingerprint combines architecture statistics with initialization-time response probes. Pairwise comparisons across the views estimate novelty. The fingerprint alone supplies architecture features for surrogate performance prediction and near-duplicate screening.}
\label{fig:representation}
\end{figure*}

\subsection{Three-View SNN Architecture Representation}
\label{sec:representation}
\label{sec:behavioral}
\label{sec:novelty}

To guide the discovery of useful native SNN architectures, we need to estimate architectural differences among generated candidates before committing to weight training. Source-code differences alone need not reflect changes in architectural mechanisms. We therefore propose a three-view representation for SNN architectures (Fig.~\ref{fig:representation}): code describes implemented operations and dataflow; the design rationale states the intended architectural idea, following EoH's idea--code pairing~\cite{liu2024evolution}; and a numerical behavioral fingerprint records structural attributes and initialization-time responses. Together, these views provide partial evidence for comparing candidates generated in the code space.

Inspired by response-based program characterization~\cite{hildebrandt2015surrogates,zhang2026rethinking}, we probe the instantiated SNN before weight training. Probe inputs, sequence shape, initialization procedure, and seed are fixed across candidate comparisons. We adapt SWAP's sample-wise pattern count~\cite{peng2024swap} to binary spike outputs, yielding sample-wise spiking patterns (SWSP): each recorded neuron--sequence-position response forms a pattern across probe samples, and SWSP counts the distinct patterns. Alongside SWSP, we use mean spike activity across spiking layers (\textsc{FireRate}) and five established activation- and gradient-based proxies~\cite{abdelfattah2021zerocost,mellor2021neural}, giving seven initialization-time measurements. We supplement them with four module statistics describing components and ten structural statistics describing dimension flow, computational organization, and resource allocation. We construct the \emph{behavioral fingerprint} $\mathbf x(a)$ as a 21-dimensional feature vector by concatenating these measurements. Feature definitions are given in Appendix~\ref{app:surrogate:features}.

For architectures $a_1$ and $a_2$, we adopt CodeBLEU~\cite{ren2020codebleu} and average its two comparison directions to obtain code similarity $s_{\mathrm{code}}(a_1,a_2)$. For rationale similarity $s_{\mathrm{rat}}(a_1,a_2)$, we encode each LLM-generated design rationale with the pretrained text-embedding model \texttt{text-embedding-v4}~\cite{alibabaTextEmbeddingV4} and compute cosine similarity between the resulting vectors. Fingerprint similarity $s_{\mathrm{fp}}(a_1,a_2)$ uses cosine similarity after fixed feature-wise affine scaling of $\mathbf x(a_1)$ and $\mathbf x(a_2)$ (Appendix~\ref{app:three-view-comparison}). We define pairwise dissimilarity using fixed, untuned equal weights:
\begin{equation}
\begin{aligned}
d(a_1,a_2)=1-\tfrac13\bigl[&
s_{\mathrm{code}}(a_1,a_2)+s_{\mathrm{rat}}(a_1,a_2)\\
&+s_{\mathrm{fp}}(a_1,a_2)\bigr].
\end{aligned}
\label{eq:distance}
\end{equation}
Inspired by novelty search~\cite{lehman2011novelty}, we estimate novelty as the mean dissimilarity to all other reference architectures:
\begin{equation}
\mathcal N(a;\mathcal R)=
\frac{\sum_{a'\in\mathcal R\setminus\{a\}}d(a,a')}
{|\mathcal R\setminus\{a\}|},\quad |\mathcal R\setminus\{a\}|>0.
\label{eq:novelty}
\end{equation}
The surrogate in Section~\ref{sec:evolution} uses evaluated architectures' fingerprints and measured task outcomes to predict task performance from an untrained candidate's $\mathbf x(a)$.

\subsection{Surrogate-Assisted Evolutionary Search}
\label{sec:evolution}
\label{sec:framework}

To limit costly weight training, we use an inner loop for surrogate-guided architecture evolution and an outer loop for training allocation and archive updates, i.e., surrogate model management~\cite{jin2011surrogate,zhang2010expensive}. Let $\mathcal D_0$ denote the initial archive of evaluated expert architectures, and $\mathcal D_t$ the archive after outer iteration $t$. Each archived architecture has an associated fingerprint and measured task outcomes. We adopt TabPFN-2.5~\cite{hollmann2025tabpfn,grinsztajn2025tabpfn} as the surrogate, using these records from $\mathcal D_{t-1}$ as labeled context. Given $\mathbf x(a)$, the surrogate predicts task metrics without training candidate $a$. The same fitness mapping used for measured outcomes aggregates these predictions into $\hat y_t(a)$; hereafter, $y(a)$ denotes fitness measured after the prescribed training protocol.

We initialize the inner-loop populations from $\mathcal D_{t-1}$ using NSGA-II's nondominated sorting and crowding-distance truncation~\cite{deb2002fast}, based on measured fitness $y(a)$ and novelty $\mathcal N(a;\mathcal D_{t-1})$. The selected architectures seed separate island populations, from which parents are sampled within or across islands. The LLM revises or recombines their code and design rationales. Feasible offspring pass fingerprint-based near-duplicate screening before surrogate evaluation. Predicted fitness guides subsequent parent selection and survival within islands, while accepted candidates accumulate in a pool for possible training.

Before selecting architectures for training, we shortlist candidates to avoid comparing every pair of generated architectures. We retain a subset prioritized by predicted performance and supplement it with candidates having high mean three-view dissimilarity to that subset. After screening against the evaluated archive for near-duplicates, we apply the same selection procedure using predicted fitness and novelty relative to the filtered pool, which remains fixed throughout selection. The resulting batch contains previously unevaluated architectures, limited by the maximum batch size $K$ and remaining budget. The selected architectures are trained under the prescribed task protocols and added to $\mathcal D_t$ with their measured outcomes, supplying additional surrogate context and parent candidates for the next iteration. Across the prescribed iterations, at most $B$ additional architectures are evaluated under these protocols. Search returns the evaluated archive and its nondominated subset under measured fitness and novelty relative to the final archive. Appendix~\ref{app:algorithms} gives the complete procedures and budget constraints.

\section{Experiments}
\label{sec:expr}

\begin{table*}[t]
\centering
\caption{\textbf{Sequence-modeling performance and estimated energy reduction.} WT103: WikiText-103 language modeling~\cite{merity2017pointer}; LRA: Long Range Arena~\cite{tay2021long}. Energy reduction is the estimated arithmetic energy of a dense Transformer of the same size divided by that of each SNN under a common architecture setting (Section~\ref{sec:main} and Appendix~\ref{app:energy}). Bold marks the best result in each column; ``--'' denotes unavailable results.}\label{tab:snn}
\footnotesize
\setlength{\tabcolsep}{4pt}
\renewcommand{\arraystretch}{1.1}
\begin{tabular}{lccccccccc}
\toprule
\multirow{2}{*}{\textbf{Model}} & \textbf{WT103} & \multicolumn{7}{c}{\textbf{LRA accuracy (\%)$\uparrow$}} & \multirow{2}{*}{\shortstack{\textbf{Energy}\\\textbf{reduction}$\uparrow$}} \\
\cmidrule(lr){2-2}\cmidrule(lr){3-9}
 & PPL$\downarrow$ & ListOps & Text & Retrieval & Image & Pathfinder & Path-X & Avg. & \\
\midrule
\multicolumn{10}{c}{\textit{ANN baselines}} \\
DeltaNet~\cite{yang2024deltanet} & 27.5 & \textbf{62.2} & 85.1 & 91.3 & 89.8 & \textbf{94.2} & 94.9 & \textbf{86.2} & -- \\
S4D-Lin$^\dagger$~\cite{gu2022s4d} & -- & 60.5 & \textbf{87.0} & 91.0 & 87.9 & 94.0 & 92.8$^*$ & 85.5 & -- \\
\midrule
\multicolumn{10}{c}{\textit{SNN baselines}} \\
S6-based SNN$^\dagger$ & -- & 55.7 & 77.6 & 88.5 & 80.1 & 83.4 & -- & -- & -- \\
Dyn-SSM$^\dagger$~\cite{zhong2024spike} & 33.2 & 60.2 & 82.4 & 88.8 & 87.2 & 92.0 & 94.4 & 84.2 & 33.1$\times$ \\
SpikingDeltaNet & 34.5 & 57.3 & 82.2 & 90.5 & 89.0 & 91.2 & 92.3 & 83.7 & 24.7$\times$ \\
SpikingMamba2 & 38.2 & 48.5 & 74.0 & 81.0 & 83.2 & 90.4 & 89.6 & 77.8 & 14.8$\times$ \\
\midrule
\rowcolor{gray!15}
\multicolumn{10}{c}{\textit{Discovered SNNs (\methodname)}} \\
\rowcolor{gray!15}
\textbf{NeuroGate} & \textbf{26.4} & 61.8 & 82.4 & 90.1 & \textbf{91.2} & 93.6 & \textbf{95.3} & 85.7 & 31.7$\times$\\
\rowcolor{gray!15}
\textbf{HomeoResSSM} & 28.1 & 46.2 & 84.8 & \textbf{92.3} & 87.6 & 86.3 & 85.2 & 80.4 & 17.3$\times$\\
\rowcolor{gray!15}
\textbf{LoopMem} & 27.8 & 59.5 & 81.0 & 89.5 & 90.5 & 88.4 & 88.7 & 82.9 & \textbf{50.6}$\times$\\
\bottomrule
\multicolumn{10}{@{}l}{\footnotesize $^\dagger$Task results quoted from~\citet{shen2025spikingssms,zhong2024spike}. $^*$S4D-Inv result.}
\end{tabular}
\end{table*}

We evaluate the discovered SNNs, examine their architectural mechanisms, and analyze how representation and search design contribute to the discovery.

\subsection{Experimental Setup}
\label{sec:setup}

Candidate blocks share the sequence-model wrapper described in Appendix~\ref{app:training-details}. During the main search, each selected candidate is trained separately for 30 epochs on WikiText-2 (WT2) language modeling and 25 epochs on one-third of the ListOps training set~\cite{merity2017pointer,tay2021long}, producing distinct model weights for the two tasks. We instantiate the composite fitness as
\begin{equation}
y(a)=\tfrac12\phi_{\mathrm{WT2}}\!\left(\mathrm{PPL}_{\mathrm{WT2}}(a)\right)
+\tfrac12\phi_{\mathrm{ListOps}}\!\left(\mathrm{Acc}_{\mathrm{ListOps}}(a)\right),
\label{eq:fitness}
\end{equation}
where the task transforms use SpikingDeltaNet as the $0.500$ fitness anchor (Appendix~\ref{app:fitness_details}). %

\begin{wrapfigure}{r}{0.5\textwidth}
\centering
\includegraphics[width=\linewidth]{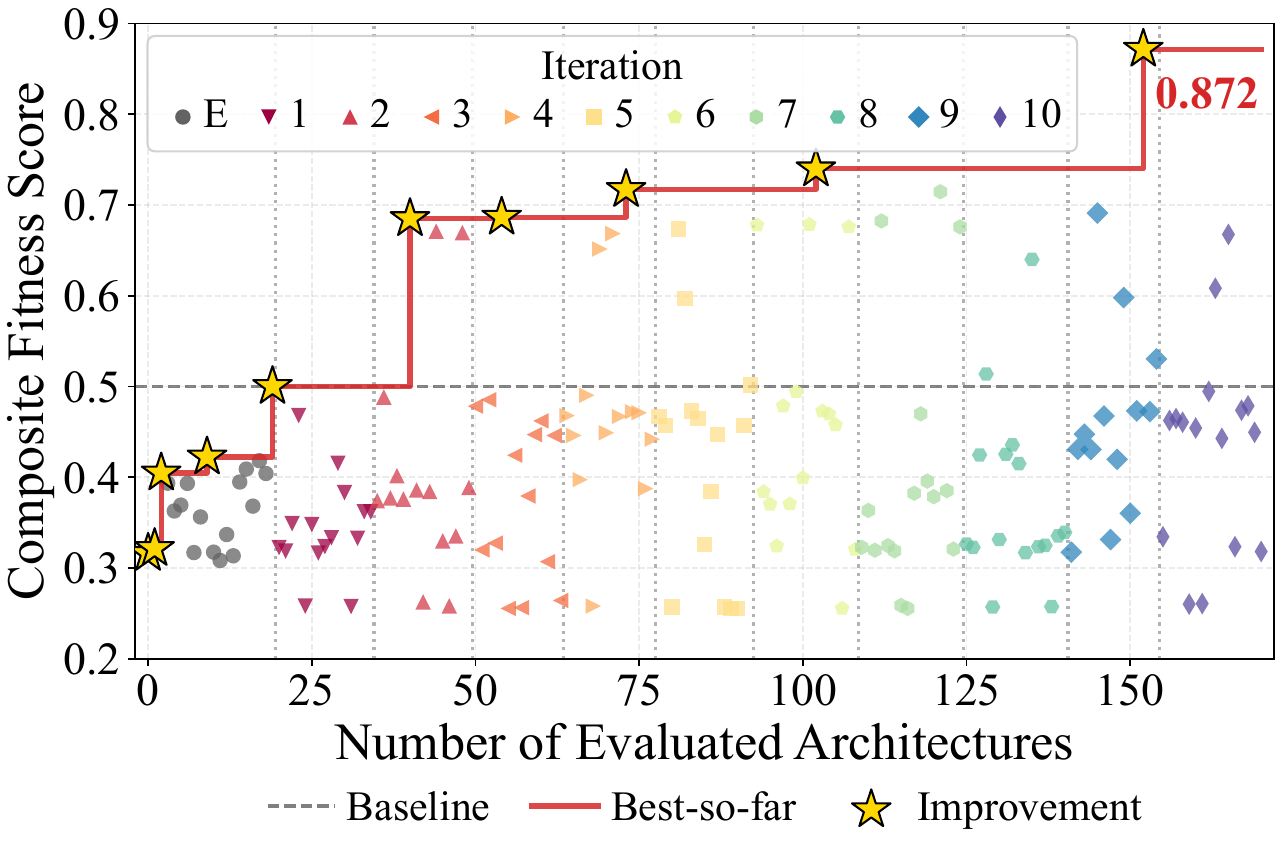}
\caption{Measured fitness of trained candidates during the main search, in evaluation order. E denotes the initial expert models (grey) and 1--10 the search iterations. The red trace shows the best-so-far fitness, including the experts, and stars mark its improvements. The dashed line marks the baseline fitness of $0.500$, attained by SpikingDeltaNet, the strongest expert.}
\label{fig:main}
\end{wrapfigure}
The initial archive contains 20 expert SNN adaptations of linear recurrent and state-space architectures (Appendix~\ref{app:zoo}). Gemini-2.5-Flash~\cite{comanici2025gemini}, selected after a comparison of candidate LLMs (Appendix~\ref{app:llm}), is used as the LLM to generate candidates, with prompts in Appendix~\ref{app:prompts}, and the surrogate specified in Section~\ref{sec:evolution} predicts their task metrics. The main search runs for 10 outer iterations with a batch limit of $K=16$ and a budget of $B=160$ architecture evaluations, each comprising both task-training runs; candidates that duplicate archived architectures are removed, so 151 new architectures are trained. Candidate weight training costs an estimated 132 V100 GPU-days (3,171 GPU-hours). Appendix~\ref{app:search} gives the search configuration and resource accounting. The code, generated programs, and evaluated archive will be released.

To evaluate the selected architectures at full scale, we retrain them on WikiText-103 (WT103)~\cite{merity2017pointer} and the six Long Range Arena (LRA) tasks, whose sequences span 1K--16K tokens~\cite{tay2021long}, following the task-specific configurations of SpikingSSMs~\cite{shen2025spikingssms}; the WT103 configuration matches that of Dyn-SSM~\cite{zhong2024spike}. Unlike the search-time setting, ListOps is trained on the full training split and evaluated on the same test split, whereas WT103 and the other five LRA tasks are not used during search (Appendix~\ref{app:training-details}). The LRA average covers all six tasks. The ANN reference, DeltaNet, is among the strongest of the 20 experts and 97 ASI-Arch models trained under our common protocol. S4D-Lin and S6-based SNN results are quoted from~\citet{shen2025spikingssms} and Dyn-SSM results from~\citet{zhong2024spike}; all other models, including our spiking adaptations SpikingDeltaNet and SpikingMamba2, are trained under our protocol. The main search comprises one run; Sections~\ref{sec:method-comparison} and~\ref{sec:ablation} report repeated smaller-scale studies.

\subsection{Main Results}
\label{sec:main}
\label{sec:bench}

Figure~\ref{fig:main} traces the main search. Candidates that surpass the strongest expert appear in every iteration from the second onward, and the best fitness improves in five separate iterations, reaching $0.872$ at iteration~9. Progress is therefore sustained rather than confined to an early proposal, consistent with the archive supplying better parents and surrogate context as it grows. The nondominated front also advances in the two task metrics, eventually containing candidates that surpass SpikingDeltaNet on both WT2 and ListOps (Appendix~\ref{app:pareto}).

\begin{wrapfigure}{r}{0.6\textwidth}
\centering
\includegraphics[width=\linewidth]{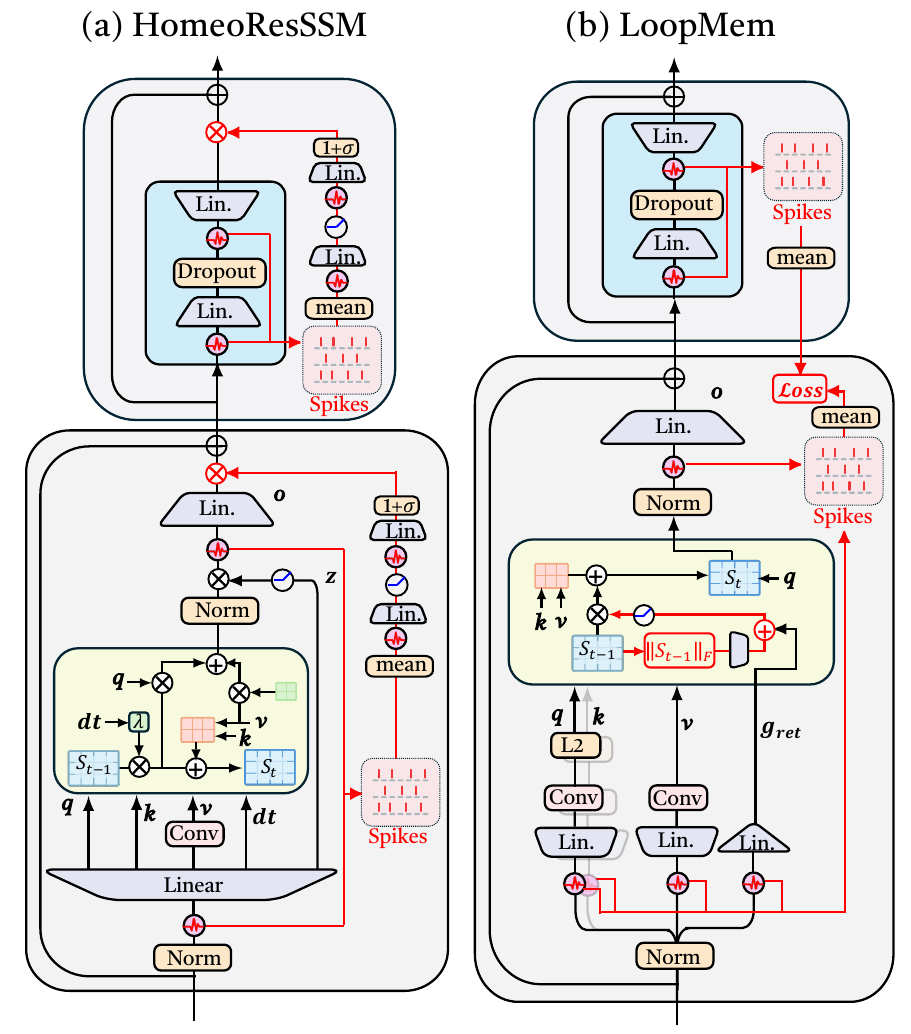}
\caption{HomeoResSSM and LoopMem. HomeoResSSM uses spike activity to modulate two residual branches. LoopMem feeds a normalized summary of its recurrent state back to the forget gate and adds activity penalties to the training loss. Red paths highlight activity and feedback connections; paths to the loss act during training. NeuroGate is shown in Fig.~\ref{fig:motivation}(c).}
\label{fig:control_relations}
\end{wrapfigure}
Using search-time measurements only, we select three architectures that represent distinct regions of the final Pareto front in composite fitness and novelty (Fig.~\ref{fig:pareto_novelty}): NeuroGate, the extreme solution with the highest fitness; HomeoResSSM, the knee point, which lies farthest from the line joining the two extreme solutions after normalization~\cite{das1999characterizing,zhang2015knee}; and LoopMem, the most novel member that retains nontrivial performance on both tasks.

The selected architectures remain strong after full-scale retraining (Table~\ref{tab:snn}). On WT103, all three outperform the listed SNN baselines, and NeuroGate also surpasses the ANN DeltaNet trained under our protocol (26.4 versus 27.5 perplexity), although its parameter-dominant projections receive binary spikes. On LRA, NeuroGate attains the highest average accuracy among the SNNs, between the two ANN baselines, and remains the highest among the SNNs without ListOps (90.5 versus 89.0), while HomeoResSSM gives the best Retrieval accuracy among all listed models. The discovered architectures thus combine competitive accuracy with distinct task profiles.

These results come with substantially lower estimated energy. Counting spike-driven projections as accumulate operations and continuous computation as multiply--accumulate operations, with each model's measured firing rate and a shared model wrapper, the discovered architectures reduce the arithmetic energy of a dense Transformer of the same size by $17$--$51\times$ (Appendix~\ref{app:energy}, which also details the Dyn-SSM estimate). The spiking feed-forward network (FFN) and output head of the shared wrapper account for most spike-driven operations, so differences among the SNNs mainly reflect their firing rates and token-mixer arithmetic. NeuroGate fires about half as often as SpikingDeltaNet, so its additional control computation still yields a larger reduction, and LoopMem attains the largest reduction among all listed models. We next examine which evolved elements distinguish the discovered architectures and whether their effects depend on spiking computation.

\subsection{Spike-Native Mechanisms in the Discovered Architectures}
\label{sec:elite-analysis}

The discovered architectures mainly contain two kinds of elements that we regard as native to SNNs. \ding{172}~\textbf{Spike-activity-dependent computation} takes spike activity itself as an input. NeuroGate maps the mean input spike activity of its query, key, and value projections at each position to gains on the delta-rule update coefficient and the output gate (Fig.~\ref{fig:motivation}(c)), and HomeoResSSM uses the spike activity of its state-space and channel-mixing branches to scale their residual contributions (Fig.~\ref{fig:control_relations}(a)); we test this class on HomeoResSSM. Because each neuron integrates its input across tokens and resets on firing, this activity reflects membrane dynamics that have no counterpart in ANN units, so the control reads a signal that exists only in spiking units. \ding{173}~\textbf{Spiking-specific design choices} do not read spikes explicitly, but depart from ANN practice in either direction: components that are uncommon in ANN design are added when they benefit the SNN, and components that are standard in ANNs are removed when they degrade it. Direct ANN-to-SNN adaptation introduces neither kind of change. LoopMem illustrates the first direction, feeding a normalized summary of its recurrent state back to the forget gate (state-norm feedback) (Fig.~\ref{fig:control_relations}(b)); NeuroGate illustrates the second, omitting the sigmoid linear unit (SiLU) activation that DeltaNet applies after its query, key, and value convolutions.

\begin{wraptable}[21]{r}{0.49\textwidth}
\centering
\caption{\textbf{Ablations of evolved elements on WT2.} Block~\ding{172} tests whether a control relies on spike activity; block~\ding{173} tests design choices that depart from ANN practice. Non-spiking variants remove the spiking neurons and all spike-dependent components. $\Delta$ is relative to the closest row above marked ``--''.}
\label{tab:elite-deltas}
\footnotesize
\setlength{\tabcolsep}{5pt}
\renewcommand{\arraystretch}{0.98}
\begin{tabular}{@{}lcc@{}}
\toprule
Variant & WT2 PPL$\downarrow$ & $\Delta$ \\
\midrule
\multicolumn{3}{@{}l}{\ding{172}~\textit{Spike-activity-dependent computation}}\\
\multicolumn{3}{@{}l}{\textbf{HomeoResSSM}}\\
\quad Discovered                          & 57.5 & -- \\
\quad Control driven by continuous inputs & 59.2 & $+1.7$ \\
\quad Control removed                     & 64.4 & $+6.9$ \\
\midrule
\multicolumn{3}{@{}l}{\ding{173}~\textit{Spiking-specific design choices}}\\
\multicolumn{3}{@{}l}{\textbf{NeuroGate}}\\
\quad Discovered (SiLU removed)           & 57.2 & -- \\
\quad SiLU restored                       & 63.1 & $+5.9$ \\
\multicolumn{3}{@{}l}{\textbf{LoopMem}}\\
\quad Discovered                          & 60.5 & -- \\
\quad State-norm feedback removed         & 69.3 & $+8.8$ \\
\cmidrule(l{1em}){1-3}
\quad Non-spiking, with feedback          & 61.3 & -- \\
\quad Non-spiking, feedback removed       & 59.7 & $-1.6$ \\
\bottomrule
\end{tabular}
\end{wraptable}
Table~\ref{tab:elite-deltas} examines these elements on WT2. All variants, including the discovered architectures, are trained with the fixed protocol applied to every search candidate, and we compare the effect of each element within an implementation rather than perplexities across implementations. For \ding{172}, removing HomeoResSSM's control raises perplexity by 6.9. Driving the same controller with the channel mean of the continuous inputs, from which the spikes are generated, recovers most of this benefit, yet the spike-driven control still attains 1.7 lower perplexity. For \ding{173}, restoring SiLU raises NeuroGate's perplexity by 5.9. Removing LoopMem's state-norm feedback raises perplexity by 8.8 in the SNN, whereas the same removal lowers the perplexity of the non-spiking variant by 1.6. LoopMem's feedback thus has opposite effects in the two implementations, and the ANN-standard SiLU degrades the SNN.

\textbf{From the search-space perspective}, each control element combines several decisions: which signal to summarize, how to transform it, and which update, branch, or gate it modulates. In \methodname, the LLM can propose these decisions within one code revision of a parent program. Among seven published encodings, covering SNN cell and block spaces, recurrent-cell spaces, and a recursive architecture grammar, none can select a spike-statistic or state-summary input, a learned controller, and its target together; directly representing these elements requires extending the encoding (Appendix~\ref{app:encoding-scope}, Table~\ref{tab:encoding-scope}). Open-code search can express such relations, but the best architecture of our adapted ASI-Arch run, SpikingCondFuse, conditions its gate on hidden-state statistics rather than spike activity (Appendix~\ref{app:asi}). Appendix~\ref{app:elite} gives the equations of the evolved elements.\looseness=-1

\subsection{Comparison with Existing Search Methods}
\label{sec:method-comparison}
\begin{figure*}[t]
\centering
\includegraphics[width=0.95\textwidth]{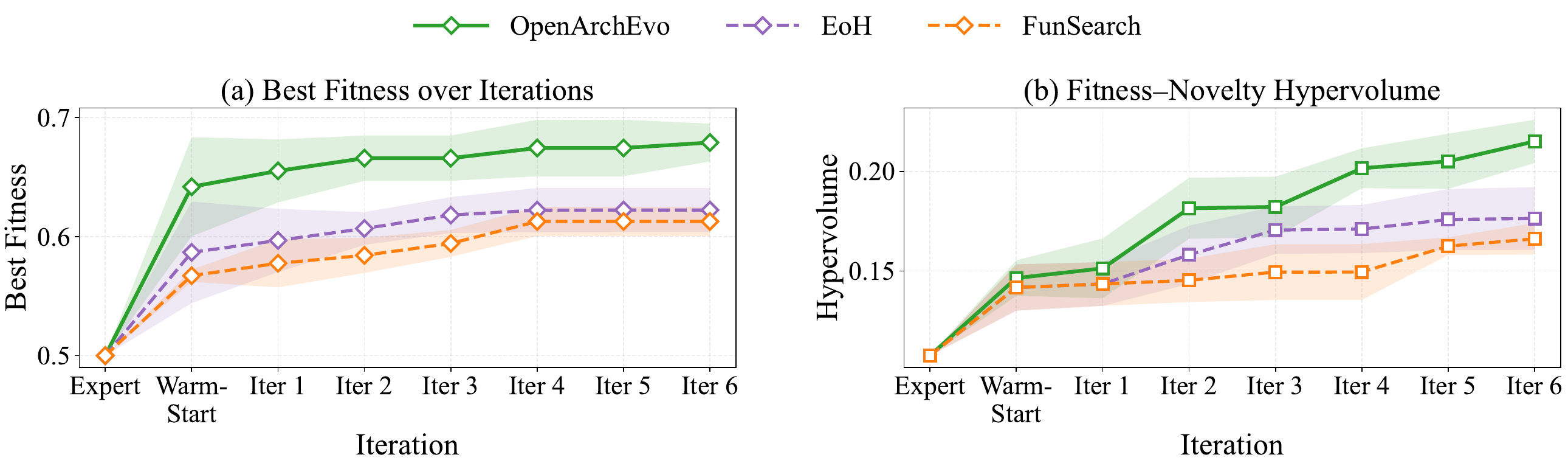}
\caption{Inner-loop search comparison in six-iteration WT2 searches: \textbf{(a) best fitness} and \textbf{(b) hypervolume} of the nondominated set in fitness and novelty after each stage. The EoH and FunSearch adaptations replace our inner-loop search and share the outer evaluation and surrogate-selection loop. Curves show means over three runs; shaded regions indicate standard deviation.}
\label{fig:ablation}
\end{figure*}
\begin{wraptable}{r}{0.5\textwidth}
\centering
\caption{Outcomes and organization of the completed program-search runs.
Downstream results compare SpikingCondFuse and NeuroGate, respectively.}
\label{tab:search_cost}
\footnotesize
\setlength{\tabcolsep}{3pt}
\renewcommand{\arraystretch}{1.10}
\begin{tabularx}{\linewidth}{@{}Xcc@{}}
\toprule
 & Adapted ASI-Arch & \methodname \\
\midrule
Best fitness$\uparrow$ & 0.653 & \textbf{0.872} \\
Top-10 mean fitness$\uparrow$ & 0.570 & \textbf{0.714} \\
WT103 PPL$\downarrow$ & 29.6 & \textbf{26.4} \\
ListOps accuracy$\uparrow$ & 55.3 & \textbf{61.8} \\
Text accuracy$\uparrow$ & 82.3 & \textbf{82.4 }\\
Retrieval accuracy$\uparrow$ & 85.3 & \textbf{90.1} \\
Energy reduction$^\ddagger$ & 18.9$\times$ & \textbf{31.7}$\times$ \\
\midrule
LLM agents & 9 & 1 \\
Prompt tokens$^\dagger$ & $\sim$25K & $\sim$1.2K \\
API expenditure (USD) & $\sim$440 & $\sim$170 \\
\bottomrule
\end{tabularx}
\vspace{2pt}
\parbox{\linewidth}{\footnotesize
One search run per method; training budgets differ as stated in the text.
$^\dagger$Prompt tokens per executable program.
$^\ddagger$Computed as in Table~\ref{tab:snn}.}
\end{wraptable}

\textbf{Inner-loop comparisons.}
We compare our inner-loop search with adaptations of EoH~\cite{liu2024evolution}
and FunSearch~\cite{romera2024mathematical}, which replace the inner loop while sharing our outer evaluation and surrogate-selection loop. Their original designs evaluate every generated program, which is impractical when each evaluation requires training an architecture. The EoH adaptation uses thought--code evolution in a single population, whereas the FunSearch adaptation uses code-based best-shot prompting with islands. Our inner loop features population management and redundancy control guided by the three-view representation and novelty.
In the main search, the trained expert archive supplies the initial surrogate context and parents. Each WT2-only search instead begins with a warm-start stage that trains a first batch of diversified candidates. Each configuration then runs six outer iterations of WT2-only search, scored by a WT2-only fitness of the same form (Appendices~\ref{app:fitness_details} and~\ref{app:asi}). \methodname\ attains the highest final best fitness and hypervolume (Fig.~\ref{fig:ablation}). All configurations use the same total training budget.

\textbf{End-to-end program-search comparison.}
We retain the agent workflow of ASI-Arch~\cite{liu2025alphago}, adapt its
evaluation tasks, and impose the spiking constraints. Table~\ref{tab:search_cost}
reports the resulting search outcomes and representative architectures.
The ASI-Arch and \methodname\ runs train 39 and 151 architectures over
six and four weeks, respectively. \methodname\ uses one program generator;
the API expenditure excludes candidate-training compute.
The best and Top-10 mean fitness are higher in our completed run.
Appendix~\ref{app:asi} provides the adaptation details.

\needspace{6\baselineskip}
\subsection{Search Analysis and Ablation Studies}
\label{sec:ablation}
\label{sec:selection_results}

\begin{wrapfigure}{r}{0.45\textwidth}
\centering
\includegraphics[width=0.94\linewidth]{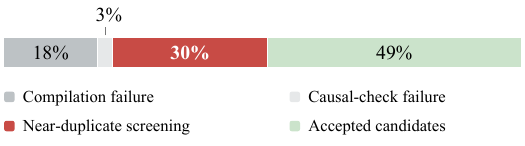}
\caption{Screening raw proposals in the main search. Compilation failures and causal-check failures account for 18\% and 3\%; fingerprint-based near-duplicate screening removes 30\%, leaving 49\% accepted. Rounded percentages share the raw-proposal denominator.}
\label{fig:duplication}
\end{wrapfigure}
\textbf{Near-duplicate screening.} Fingerprint-based near-duplicate screening removes 30\% of raw proposals (Fig.~\ref{fig:duplication}). The shared initialization-time probes and architecture statistics identify these candidates before surrogate-guided selection.

\begin{figure*}[t]
\centering
\includegraphics[width=0.98\textwidth]{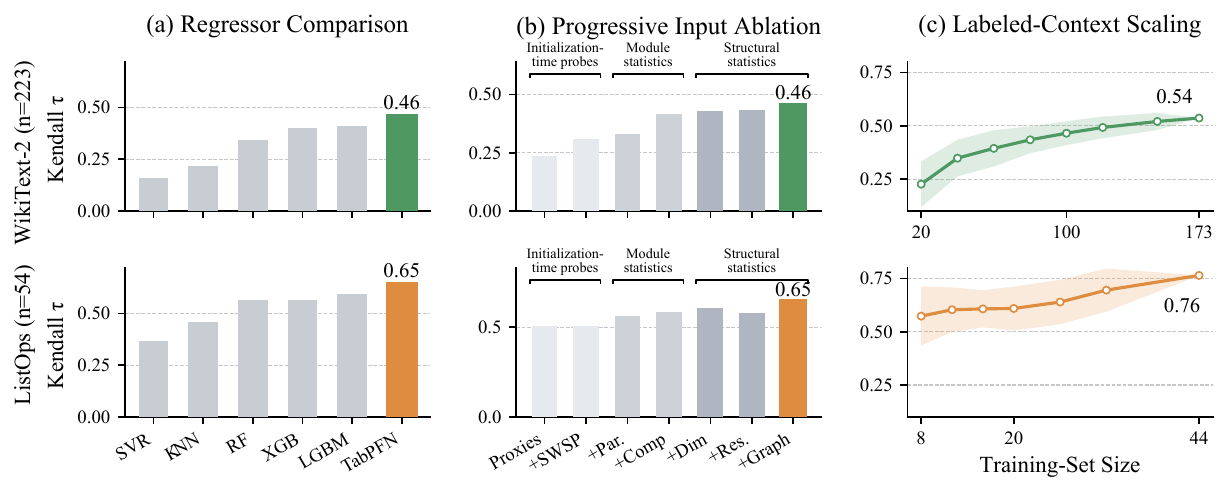}
\caption{Offline surrogate studies on WT2 (top, 223 architectures) and ListOps (bottom, 54). Kendall $\tau$ compares regressors (a), progressive surrogate-input additions (b), and labeled-context sizes (c). The first two studies share repeated cross-validation splits. In (b), gray shades mark the measurement group added at each cumulative step, and the task color marks the complete fingerprint. The +SWSP step adds both SWSP and FireRate; +Par.\ adds the parameter count and +Comp the other module statistics, while +Dim, +Res., and +Graph add the dimension-flow, resource-allocation, and subgraph statistics (Table~\ref{tab:features}). Learning curves use fixed holdouts and 20 training-pool subsamples per size, with $\pm1\sigma$ bands. Appendix~\ref{app:surrogate} gives the protocols.}
\label{fig:surrogate_composite}
\end{figure*}

\textbf{Surrogate prediction.} The surrogate studies were conducted before the main search and guided the choice of its surrogate and fingerprint (Appendix~\ref{app:surrogate}). Initialization-time probes capture network responses and form the starting point for our surrogate-input study (Fig.~\ref{fig:surrogate_composite}(b)). Adding SWSP and firing rate to the classical proxies improves ranking on WT2, with a smaller change on ListOps. Supplementing these probes with module and structural statistics gives the strongest ranking along the tested sequence. With the complete fingerprint, TabPFN-2.5 ranks best among the tested regressors on both tasks (Fig.~\ref{fig:surrogate_composite}(a)). In these studies, prediction generally improves as the labeled context grows (Fig.~\ref{fig:surrogate_composite}(c)). Early in search, candidates are selected for training by a surrogate with little labeled context. In WT2-only searches without the warm-start stage, which start from the expert archive alone, the best fitness after six iterations is 0.56 instead of 0.68, and the gap persists across iterations (Appendix~\ref{app:asi}).

\begin{wraptable}{r}{0.6\textwidth}
\centering
\caption{Selection variants in six-iteration WT2 searches (three runs). Statistics are computed over the final-iteration trained candidates pooled across runs: Top-1 and Top-5 give the lowest and the mean of the five lowest PPL, Novelty the mean novelty, and Variance the variance of WT2 fitness.}\label{tab:exp3}
\rowcolors{2}{gray!8}{white}
\footnotesize
\setlength{\tabcolsep}{3pt}
\begin{tabular}{@{}lcccc@{}}
\toprule
\textbf{Selection method} & \shortstack{Top-1\\PPL$\downarrow$} & \shortstack{Top-5\\PPL$\downarrow$} & Novelty$\uparrow$ & Variance$\downarrow$ \\
\midrule
Evolution + surrogate           & \textbf{56.2} & \textbf{57.4} & \textbf{0.534} & 0.0281 \\    
Single island + surrogate & 57.3 & 57.9 & 0.513 & \textbf{0.0272} \\
Sampling + surrogate   & 57.3 & 57.8 & 0.520 & 0.0314 \\
Sampling only      & 57.5 & 58.5 & 0.511 & 0.0414 \\
\bottomrule
\end{tabular}
\end{wraptable}
\textbf{Search ablations.} We compare candidate-generation and selection variants in six-iteration WT2-only searches, each repeated three times. Besides the full configuration, which evolves candidates on multiple islands and selects them with the surrogate, the variants evolve a single island, sample candidates from the LLM without evolution before surrogate selection, or sample and select without the surrogate. The full configuration has the lowest Top-1 and Top-5 PPL and the highest novelty, while the single-island variant has the lowest fitness variance (Table~\ref{tab:exp3}).

\section{Discussion}
\label{sec:discussion}

Program evolution discovers changes in control and pathway design around established recurrent operators, including how spike activity regulates computation. It shifts part of architecture design from enumerating admissible mechanisms to specifying executable constraints and evaluating proposed programs. The resulting archive contains both usable architectures and design hypotheses for further study.

For expensive evolutionary search, the framework separates abundant program variation from limited training evaluations. A shared characterization supports redundancy filtering, population organization, and surrogate prediction, allowing these components to improve together. The evidence comprises one full search with smaller repeated studies; the ASI-Arch comparison uses unequal budgets. Broader validation across search runs and domains remains necessary, as fingerprints approximate redundancy and novelty depends on the reference population.

The discovered SNNs demonstrate that program evolution can identify competitive spiking sequence architectures with spike-activity-dependent computation and spiking-specific design choices. Their energy reductions are theoretical arithmetic estimates; realizing and measuring these reductions requires implementations on specific hardware that account for memory access and execution costs. The present search fixes the neuron model. Extending it to jointly evolve neuron dynamics and architecture, and evaluating broader and more demanding downstream tasks, are natural next steps. These results show that program search can discover competitive SNN architectures, which motivates extending it to those harder settings.

\section{Conclusion}
We introduced \methodname\ for automated discovery of native neural architectures, with spiking sequence modeling as a demanding test case. The discovered architectures combine competitive predictive performance with substantial estimated arithmetic-energy savings, demonstrating the value of exploring computations tailored to spike activity. Beyond the resulting models, the study shows how executable architectural proposals can become a cumulative record of empirically evaluated designs. The broader opportunity is to connect expressive program generation with meaningful architectural comparison and selective evaluation, enabling discovery in neural computing domains where useful mechanisms are difficult to specify in advance.

\section*{Acknowledgments}
Large language models were used to polish the language of this manuscript; the authors take full responsibility for its content.

{\small\putbib}
\end{bibunit}

\clearpage
\appendix
\renewcommand{\topfraction}{0.9}\renewcommand{\bottomfraction}{0.6}\renewcommand{\textfraction}{0.1}\renewcommand{\floatpagefraction}{0.8}
\begin{bibunit}
\def\HyCitePfx{app.}
\section{Supporting Architecture Comparisons}
\subsection{Paired ANN and SNN Evaluation}
\label{app:ann_snn_mismatch}
We adapted the 106 architecture programs released by ASI-Arch~\cite{liu2025alphago} to the spiking-projection constraint below and trained the ANN and SNN implementations separately from scratch on WT2 and ListOps. Nine SNN runs produced non-finite loss; the remaining 97 architectures have complete paired results on both tasks. This pool is distinct from the 20 experts used to initialize our search. The WT2 ANN--SNN comparison uses test perplexity and tie-adjusted Kendall $\tau_b$, giving $0.2997$ on the 97 successful pairs.

\label{app:cap_ab}
On these same architecture identifiers, the cross-task rank correlation between WT2 perplexity and ListOps accuracy is $0.0516$ for SNNs and $0.3031$ for ANNs, with lower perplexity and higher accuracy oriented as better.

\subsection{Scope of the Spiking Constraint}
\label{app:fully_snn_definition}
The spiking-projection constraint places binary spike inputs before the parameter-dominant feature projections. Membrane potentials, residual streams, recurrent states, and token-mixer arithmetic remain continuous. Lightweight control computations may also be continuous, including Comba's closed-loop state feedback and LoopMem's evolved state-norm feedback network. Thus the constraint does not imply that all arithmetic is spike-driven. Candidate blocks use the shared hard-reset LIF implementation in Eq.~\ref{eq:lif}; the surrogate dynamic network (SDN) of SpikingSSMs~\cite{shen2025spikingssms} accelerates the neuron computation during search and evaluation. Candidates must also preserve the block interface, causality, and subquadratic sequence complexity.

\section{Architectural Changes and the Scope of Published Search Encodings}
\label{app:encoding-scope}

We compare the dependencies in the discovered blocks with the choices exposed by published search encodings. Table~\ref{tab:encoding-scope} records the available primitives and composition rules, together with the extensions needed to directly represent the relevant computation. This is an encoding-level comparison: the methods address different tasks and use different evaluation protocols. The criterion is whether a specified computational relation is selectable under the documented encoding, rather than whether another network could approximate its input--output function.

\begin{table}[h]
\centering
\caption{Scope of documented search encodings. The final column identifies missing choices for directly representing the examined SNN computations. Source locations specify the versions and sections inspected; extending an encoding can change its scope.}
\label{tab:encoding-scope}
\footnotesize
\setlength{\tabcolsep}{5pt}
\renewcommand{\arraystretch}{1.16}
\begin{tabularx}{\textwidth}{@{}p{0.19\textwidth} X X@{}}
\toprule
Method and source location & Encoded architectural choices & Relation to the discovered computations \\
\midrule
AutoSNN~\cite{na2022autosnn}, Sec.\,4.1 & Five block choices in a fixed backbone: skip, spiking convolution, and spiking residual blocks, with specified kernel sizes. & The block menu needs a spike-statistic controller and its target connection to select the activity-dependent update or residual modulation. \\
SNASNet~\cite{kim2022snasnet}, Cell Search Strategy & Four-node cells with zero, skip, convolution, and pooling operations on forward and cross-time backward edges. & Temporal feedback is already permitted. A spike-statistic reduction and learned multiplicative control are additional operations beyond this edge menu. \\
MSE-NAS~\cite{pan2025msenas}, Sec.\,III, Fig.\,1 & A multiscale genotype selects layer operations, excitatory/inhibitory types, motifs, and global connections under a specified decoder. & These choices alter operations and connectivity; the decoder would need to expose activity-conditioned control equations and their attachment points. \\
EQ-SpikeLM~\cite{zhang2026eqspikelm}, Sec.\,IV-B.1, Eqs.\,(14)--(16) & Per-layer preserved channel ratios for query--key, value, and feed-forward projections in a pretrained spiking language model. & Channel pruning changes widths within the existing computation; it does not introduce a new spike-derived control dependency. \\
ENAS (efficient NAS), recurrent space~\cite{pham2018enas}, Secs.\,2.1, 3.1 & Predecessor and activation choices in a recurrent cell, with a prescribed highway-gating construction. & Recurrence and multiplicative gating are present. Spike-statistic inputs and revisions to the controller's equation or target require extending the template. \\
DARTS, recurrent space~\cite{liu2019darts}, Sec.\,3.1.2 & Operation choices over linear transforms and activations, identity, and zero, within a recurrent-cell template with highway bypasses. & Selecting operations does not itself expose the spike-reduction and control-target relation; these must be added to the operations or template. \\
\textit{einspace}~\cite{ericsson2024einspace}, Secs.\,3.1--3.3, 5 & Recursive grammar for sequential, branching, routing, and computation modules, including matrix multiplication, summation, and concatenation. & Composition is substantially broader than a fixed cell. The published grammar excludes recurrent computation, so the complete recurrent SNN blocks require a grammar extension. \\
\bottomrule
\end{tabularx}
\end{table}

\textbf{The computational relations being compared.}
NeuroGate forms activity statistics from spikes and uses learned transformations to modulate the recurrent update coefficient and output-gate input. HomeoResSSM constructs separate activity-conditioned gains for the state-space and channel-mixer residual branches. Their distinguishing dependency is therefore \emph{spike statistic $\rightarrow$ learned control $\rightarrow$ a specified update or branch}, within an otherwise inherited recurrent core. LoopMem provides a related case: a normalized summary of the recurrent state controls the forget gate, giving the dependency \emph{state summary $\rightarrow$ learned control $\rightarrow$ gate}, which likewise requires a summary input and a controller attached to the gate.

\section{Search Implementation}
\label{app:algorithms}
Algorithms~\ref{alg:sandnas} and~\ref{alg:algobleu} specify the outer and inner loops. Each nondominated ranking uses a fixed reference set for novelty: the evaluated archive for parent selection and the filtered candidate pool for training-batch selection. $\operatorname{ND}_{\mathcal R}(\mathcal S)$ denotes the nondominated subset of $\mathcal S$ under $[y(a),\,\mathcal N(a;\mathcal R)]$. The new-evaluation budget excludes the initial expert archive. Island survival changes the active parent population; accepted programs remain in the cumulative candidate pool.
\RestyleAlgo{ruled}
\begin{algorithm}[t]
\small
\SetAlgoLined
\SetKwInOut{Input}{Input}\SetKwInOut{Output}{Output}
\DontPrintSemicolon
\caption{Outer loop: selective real-training allocation}
\label{alg:sandnas}
\Input{Initial evaluated archive $\mathcal{D}_0$; rounds $T$; pool target $n_{\mathrm{pool}}$; batch limit $K$; new-evaluation budget $B$}
\Output{Evaluated archive $\mathcal{D}_T$ and $\operatorname{ND}_{\mathcal D_T}(\mathcal D_T)$}
\For{$t = 1, \ldots, T$}{
    Refresh surrogate $\mathcal{S}_t$ from $\mathcal{D}_{t-1}$\;
    Form $\mathcal P_{t0}\subseteq\mathcal D_{t-1}$ of size $|\mathcal D_0|$ using NSGA-II's nondominated sorting and crowding-distance truncation on $[y(a),\,\mathcal N(a;\mathcal D_{t-1})]$\;
    $\mathcal{P}_t \leftarrow$ Algorithm~\ref{alg:algobleu}$(\mathcal{S}_t,\mathcal P_{t0})$ \tcp*{surrogate-guided variation}
    $\mathcal{P}_{\mathrm{top}} \leftarrow$ globally best candidates, the best candidates of each island, and stepping stones that raised the best-so-far $\hat y_t$, filled to the shortlist size by $\hat y_t$\;
    $\mathcal{P}_{\mathrm{dissim}} \leftarrow$ lowest mean similarity to $\mathcal{P}_{\mathrm{top}}$ among remaining candidates\;
    $\mathcal{P}_f \leftarrow \mathcal{P}_{\mathrm{top}} \cup \mathcal{P}_{\mathrm{dissim}}$\;
    $\mathcal{P}_f \leftarrow$ screen $\mathcal{P}_f$ against the evaluated archive for near-duplicates\;
    Set remaining budget $b_t\leftarrow B-|\mathcal D_{t-1}\setminus\mathcal D_0|$\;
    Select batch $\mathcal Q_t$ of at most $\min(K,b_t)$ candidates by the same rule on $[\hat y_t(a),\, \mathcal{N}(a;\mathcal{P}_f)]$\;
    Train each $a\in\mathcal Q_t$ on the proxy tasks; compute task metrics and $y(a)$\;
    $\mathcal{D}_t \leftarrow \mathcal{D}_{t-1} \cup \mathcal Q_t$\;
    Store code, rationale, fingerprint, task metrics, and fitness for each new archive member\;
}
\Return{$\mathcal{D}_T$ and $\operatorname{ND}_{\mathcal D_T}(\mathcal D_T)$}
\end{algorithm}

\begin{algorithm}[t]
\small
\SetAlgoLined
\SetKwInOut{Input}{Input}\SetKwInOut{Output}{Output}
\DontPrintSemicolon
\caption{Inner loop: surrogate-guided program evolution}
\label{alg:algobleu}
\Input{Surrogate $\mathcal{S}_t$; initial population $\mathcal P_{t0}$; target size $n_{\mathrm{pool}}$; validity predicate $\mathcal{C}$; $M$ islands of capacity $c_{\mathrm{isl}}$}
\Output{Candidate pool $\mathcal{P}_t$}
Spectrally cluster $\mathcal P_{t0}$ into $M$ islands using $\operatorname{Sim}=1-d$ (one island in the warm-start stage)\; $\mathcal{P}_t \leftarrow \emptyset$\;
\While{$|\mathcal{P}_t| < n_{\mathrm{pool}}$}{
    Sample parents within or across islands; propose program $a$ and rationale via LLM mutation\;
    \lIf{program execution fails or $\mathcal C(a)=0$}{\textbf{continue}}
    Extract fingerprint $\mathbf{x}(a)$\;
    \lIf{fingerprint nearly matches a parent}{\textbf{continue}}
    Predict task metrics using $\mathcal S_t$; compute $\hat y_t(a)$\;
    Assign $a$ to island $I_m$ of greatest mean similarity $\bar s_m(a)$\;
    \lIf{registration rejects a near-redundant proposal}{\textbf{continue}}
    Add $a$ to $I_m$ and $\mathcal{P}_t$\;
    \If{$|I_m| \geq 2c_{\mathrm{isl}}$}{retain $c_{\mathrm{isl}}$ active members by island survival\;}
    \If{reset interval has elapsed}{
        Rank islands by their best predicted fitness; clear the weaker half\;
        Seed each cleared island with the best member of a sampled retained island\;
    }
}
\Return{$\mathcal{P}_t$}
\end{algorithm}

\subsection{Fingerprint and Three-View Comparison}
\label{app:surrogate:features}
The 21 measurements in Table~\ref{tab:features} are extracted from the instantiated model with its fixed outer wrapper. Module statistics use the module inventory; structural measurements use the execution trace and tensor shapes. Initialization-time probes use a fixed ListOps minibatch, sequence shape, initialization procedure, and seed across candidates. The five classical proxies follow the corresponding definitions~\cite{tanaka2020synflow,abdelfattah2021zerocost,lee2019snip,mellor2021neural}.

For SWSP, collect binary responses in $S\in\{0,1\}^{n\times V}$, with $n$ probe samples and $V$ recorded neuron--sequence-position responses. Following the sample-wise orientation of SWAP~\cite{peng2024swap},
\begin{equation}
\mathrm{SWSP}=\left|\{S_{:,v}:v=1,\ldots,V\}\right|.
\end{equation}
FireRate is the mean spike activity across the recorded spiking layers at initialization. Both are surrogate inputs measured before training.

\begin{table}[t]
\centering
\caption{Complete 21-dimensional behavioral fingerprint used in Section~\ref{sec:novelty}. The first two groups form architecture statistics; the third comprises initialization-time probes. Type indicates extraction cost: C\,=\,code analysis (static), F\,=\,forward-based measurement, F+B\,=\,forward\,+\,backward computations.}
\label{tab:features}
\small
\setlength{\tabcolsep}{4pt}
\renewcommand{\arraystretch}{1.08}
\begin{tabular}{@{}llcl@{}}
\toprule
\textbf{Group} & \textbf{Measurement} & \textbf{Type} & \textbf{Description} \\
\midrule
\shortstack[l]{\textsc{Module}\\\textsc{statistics}}
 & Params                       & C   & Total trainable parameters \\
 & SpkLinear                    & C   & Spiking linear module count \\
 & Conv1d                       & C   & 1-D convolution count \\
 & GateRatio                    & C   & Gating layer fraction \\
\midrule
\textsc{Structural statistics}
 & Depth                        & F   & Shape-changing layer count \\
 & Expand                       & F   & Dim-increase steps \\
 & Contract                     & F   & Dim-decrease steps \\
 & BneckRatio                   & F   & Min-dim / model-dim \\
 & MaxExpand                    & F   & Largest expansion factor \\
 & FFN\textsubscript{exp}       & F   & Avg FFN expansion ratio \\
 & FFN\textsubscript{param}     & F   & FFN parameter fraction \\
 & FFN\textsubscript{mac}       & F   & FFN MAC fraction \\
 & MAC/Param                    & F   & Compute density \\
 & SubgraphR                    & F   & Largest FX subgraph ratio \\
\midrule
\shortstack[l]{\textsc{Initialization-time}\\\textsc{probes}}
 & SynFlow                      & F+B & Synaptic flow \\
 & GradNorm                     & F+B & Gradient $\ell_2$ norm \\
 & SNIP                         & F+B & Connection sensitivity \\
 & Jacob\textsubscript{cov}     & F+B & Jacobian covariance \\
 & NASWOT                       & F & Activation overlap \\
 & SWSP                         & F & Sample-wise spiking patterns \\
 & FireRate                     & F & Spiking firing rate \\
\bottomrule
\end{tabular}
\end{table}

\subsection{Three-View Similarity}
\label{app:three-view-comparison}
Code similarity averages full Python CodeBLEU in both reference--candidate directions, with equal weights on its four components~\cite{ren2020codebleu}. Rationale similarity uses cosine similarity between 2,048-dimensional \texttt{text-embedding-v4} embeddings~\cite{alibabaTextEmbeddingV4}. Before fingerprint cosine similarity, each feature is transformed by
\begin{equation}
\widetilde x_j=c\left(2\,\frac{x_j-\ell_j}{u_j-\ell_j}-1\right),
\end{equation}
which maps $[\ell_j,u_j]$ to $[-c,c]$; we set $c=0.9$. The 21 reference pairs were calibrated on the 97 ASI-Arch paired architectures, satisfy $u_j>\ell_j$, and remain fixed throughout search. The transform is applied without clipping; cosine similarity is set to zero if either vector has zero norm. The three views have equal weights as in Eq.~\ref{eq:distance}.

\subsection{Initial Expert Archive}
\label{app:zoo}
The initial archive $\mathcal D_0$ contains 20 linear recurrent and state-space architectures from five families (Table~\ref{tab:model_zoo}), each adapted to the spiking-projection constraint in Appendix~\ref{app:fully_snn_definition}. Eighteen are taken from the Flash Linear Attention library~\cite{yang2024fla}; S4~\cite{gu2022s4} and MetaLA~\cite{chou2024metala} extend coverage of state-space models and modern RNNs. The families span data-independent and data-dependent gating, delta-rule updates, and closed-loop state feedback. Designs that augment the recurrent state with an external memory hierarchy, such as the hierarchical memory for Mamba~\cite{wang2026mamba}, are not included. All 20 are trained on both proxy tasks to supply initial parents and labeled surrogate context; the spiking implementations will be released with the code.

\begin{table}[h]
\centering
\small
\setlength{\tabcolsep}{4pt}
\renewcommand{\arraystretch}{1.05}
\caption{The 20 architectures in the initial expert archive $\mathcal{D}_0$, each adapted to the spiking-projection constraint.}\label{tab:model_zoo}
\resizebox{\linewidth}{!}{%
\begin{tabular}{@{}lcll@{\hspace{18pt}}lcll@{}}
\toprule
\textbf{Model} & \textbf{Family} & \textbf{Venue} & \textbf{Ref.} & \textbf{Model} & \textbf{Family} & \textbf{Venue} & \textbf{Ref.} \\
\cmidrule(r{9pt}){1-4}\cmidrule(l){5-8}
RetNet & Foundational & arXiv 2023 & \cite{sun2023retnet} & S4 & SSM & ICLR 2022 & \cite{gu2022s4} \\
LightNet & Foundational & TMLR 2026 & \cite{qin2024lightnet} & Mamba (S6) & SSM & COLM 2024 & \cite{gu2023mamba} \\
GLA & Modern RNN & ICML 2024 & \cite{yang2024gla} & Mamba2 (SSD) & SSM & ICML 2024 & \cite{dao2024mamba2} \\
HGRN & Modern RNN & NeurIPS 2023 & \cite{qin2023hgrn} & DeltaNet & Delta Rule & NeurIPS 2024 & \cite{yang2024deltanet} \\
HGRN2 & Modern RNN & COLM 2024 & \cite{qin2024hgrn2} & DeltaFormer & Delta Rule & arXiv 2025 & \cite{zhong2025deltaformer} \\
RWKV-6 & Modern RNN & COLM 2024 & \cite{peng2024eagle} & Gated DeltaNet & Delta Rule & ICLR 2025 & \cite{yang2024gated} \\
RWKV-7 & Modern RNN & COLM 2025 & \cite{peng2025rwkv7} & KDA & Delta Rule & arXiv 2025 & \cite{kimiteam2025kimilinear} \\
ABC & Modern RNN & ACL 2022 & \cite{peng2022abc} & DeltaProduct & Delta Rule & NeurIPS 2025 & \cite{siems2025deltaproduct} \\
GSA & Modern RNN & NeurIPS 2024 & \cite{zhang2024gsa} & MesaNet & Advanced & ICLR 2026 & \cite{vonoswald2025mesanet} \\
MetaLA & Modern RNN & NeurIPS 2024 & \cite{chou2024metala} & Comba & Advanced & NeurIPS 2025 & \cite{hu2025comba} \\
\bottomrule
\end{tabular}}
\end{table}

\subsection{Search Configuration and Training Cost}
\label{app:search}
\begin{table}[h]
\centering\small
\caption{Main-search configuration.}\label{tab:hparams}
\begin{tabular}{@{}ll@{}}\toprule
Setting & Value \\
\midrule
LLM / maximum generation tokens & Gemini-2.5-Flash / 32,768 \\
Outer iterations / accepted candidates per iteration & 10 / 1,080 \\
Batch limit $K$ / evaluation budget $B$ & 16 / 160 \\
Islands / retained capacity & 10 / approximately 20 per island \\
Population trimming threshold & Twice the retained capacity \\
Inter-island parent sampling probability & 0.5 \\
Reset interval / fraction of islands reset & 3,600 seconds / one-half \\
Performance / dissimilarity shortlist sizes & 30 / 30 \\
Similarity early-exit tolerance & $10^{-6}$ \\
\bottomrule\end{tabular}
\end{table}
Each iteration selects at most $K=16$ candidates for training; candidates that duplicate archived architectures are removed, so some batches are smaller and 151 new architectures are trained in total. Each candidate-training job uses one NVIDIA V100 (32\,GB). Approximate per-architecture training costs are one GPU-hour on WT2 and 20 GPU-hours on ListOps, giving $151(1+20)=3{,}171$ GPU-hours, or about 132 GPU-days, excluding initial-expert training, downstream retraining, and LLM inference.

\needspace{20\baselineskip}
\subsection{LLM Selection}
\label{app:llm}
\begin{wrapfigure}{r}{0.55\textwidth}
\centering
\includegraphics[width=\linewidth]{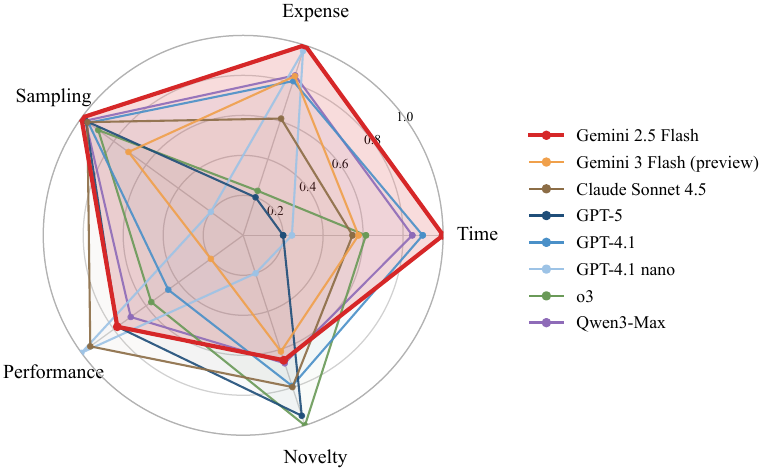}
\caption{Comparison of candidate LLMs as program generators. Each axis is normalized across models, with 1.0 for the best and 0.2 for the worst.}
\label{fig:llm}
\end{wrapfigure}
Before the main search, candidate LLMs, including Gemini-2.5-Flash~\cite{comanici2025gemini}, OpenAI o3~\cite{openai2025o3}, and Qwen3-Max~\cite{qwen2025max}, were compared as program generators on sampling success, performance and novelty of the generated architectures, generation time, and API expense (Fig.~\ref{fig:llm}). No model is best on every axis. Because the search issues more than two thousand generation calls per iteration (1,080 accepted candidates at a 49\% acceptance rate; Table~\ref{tab:hparams} and Fig.~\ref{fig:duplication}), its time and expense scale with the number of calls. Gemini-2.5-Flash is best in time, expense, and sampling success and is therefore used in all reported searches.

\subsection{Pareto-Front Evolution}
\label{app:pareto}
\begin{figure*}[!htb]
\centering
\includegraphics[width=0.94\textwidth]{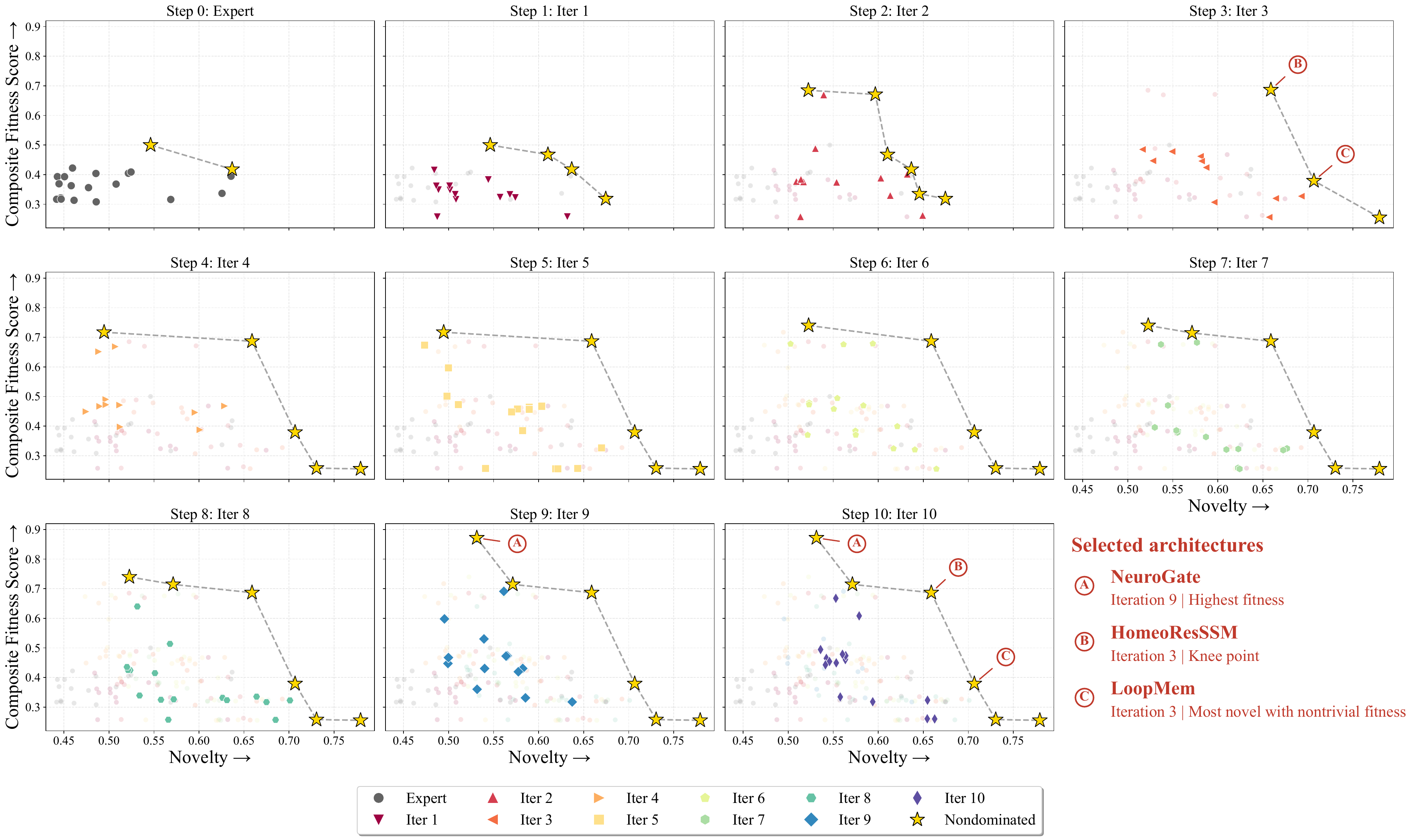}
\caption{Evolution of the nondominated front in composite fitness and novelty during the main search. Each panel adds the architectures trained in one iteration (colored) to those of earlier iterations (faded); stars mark the nondominated architectures. Step~0 contains the initial expert models. Labels A--C mark the three selected architectures in the iteration where each first appears and on the final front.}
\label{fig:pareto_novelty}
\end{figure*}

The outer loop selects candidates by two objectives, composite fitness and novelty (Section~\ref{sec:evolution}). Fig.~\ref{fig:pareto_novelty} shows the nondominated front of all trained architectures in these objectives after each iteration. The front advances along both axes. The selected architectures are marked where they first appear and on the final front: NeuroGate (iteration~9) has the highest fitness; HomeoResSSM (iteration~3) lies at the knee point, the member farthest from the line joining the two extreme solutions after min--max normalization of both objectives; and LoopMem (iteration~3) is the most novel member with nontrivial performance on both tasks, as the two more novel members reach only 18.7\% and 17.8\% ListOps accuracy.

\begin{figure*}[!htb]
\centering
\includegraphics[width=0.94\textwidth]{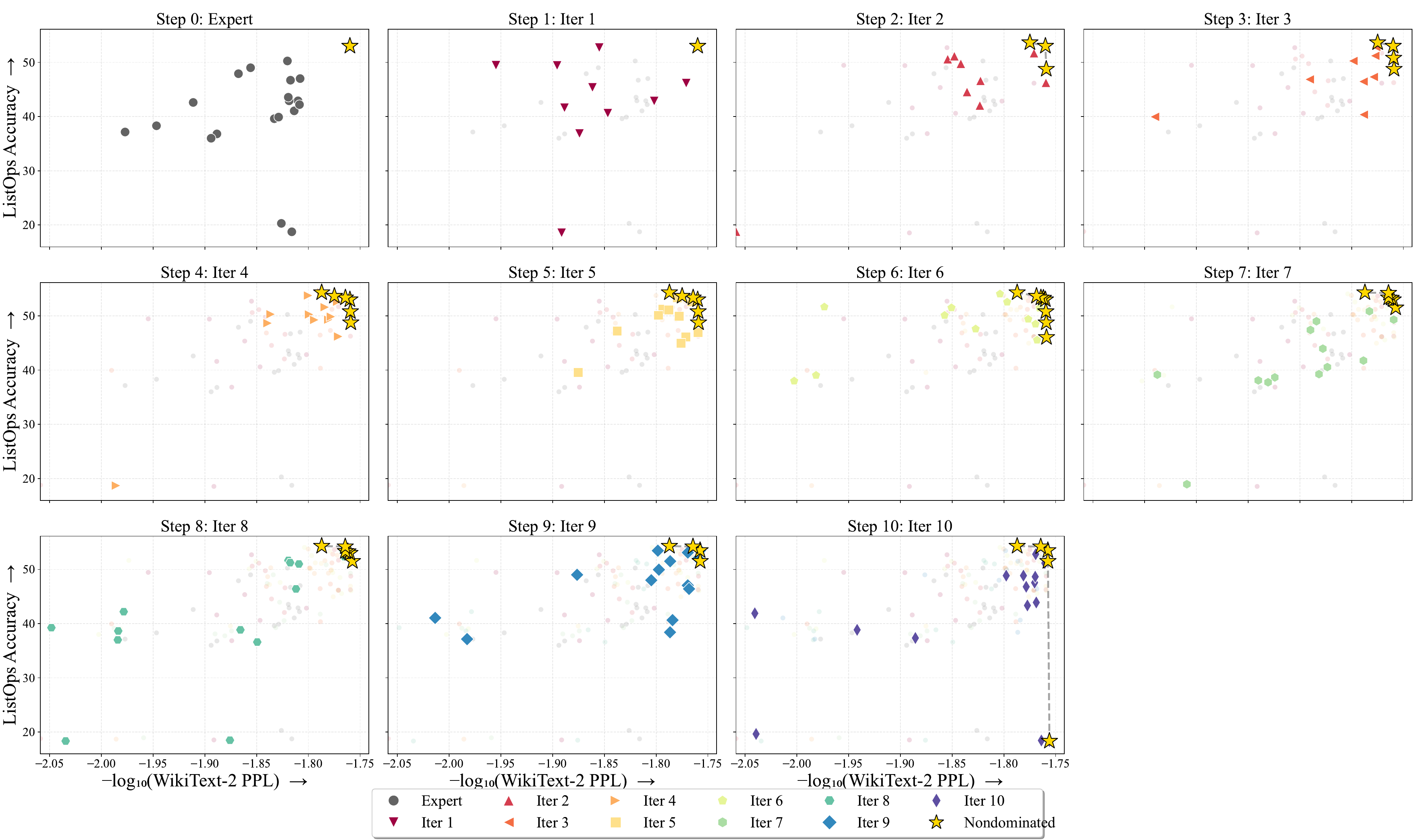}
\caption{Evolution of the nondominated front in ListOps accuracy and WT2 perplexity during the main search, displayed as in Fig.~\ref{fig:pareto_novelty}; the horizontal axis shows $\log_{10}$ WT2 perplexity, decreasing to the right.}
\label{fig:pareto_accppl}
\end{figure*}

The composite fitness itself combines two task objectives. Fig.~\ref{fig:pareto_accppl} shows the corresponding front in ListOps accuracy and WT2 perplexity. At initialization, SpikingDeltaNet alone forms this front. As the search proceeds, the front extends in both directions, and four search candidates, including the one with the highest fitness, improve on SpikingDeltaNet in both metrics.

\needspace{24\baselineskip}
\section{Evaluation Protocols}
\subsection{Training Tasks and Fixed Model Components}
\label{app:training-details}
\begin{wraptable}{r}{0.45\textwidth}
\caption{Proxy-training configurations used to compute search fitness.}
\label{tab:training_main}
\centering\small
\begin{tabular}{@{}lcc@{}}
\toprule
Setting & WT2 & ListOps \\
\midrule
Block layers / width & 6 / 256 & 2 / 128 \\
Attention heads & 8 & 4 \\
FFN expansion & 4 & 2 \\
Dropout & 0.2 & 0.1 \\
Batch size & 16 & 32 \\
Epochs & 30 & 25 \\
Learning rate & $6\times10^{-3}$ & $10^{-3}$ \\
Weight decay & 0.15 & $10^{-4}$ \\
Warmup steps & 600 & 3000 \\
Gradient clipping norm & 2.0 & 2.0 \\
Optimizer & AdamW & AdamW \\
Schedule & Cosine & Cosine \\
\bottomrule
\end{tabular}
\end{wraptable}
Search-time WT2 fitness uses validation perplexity. WT2 uses GPT-2 byte-level BPE, vocabulary size 50,257, and concatenated token streams chunked into length-512 sequences. ListOps uses a training-derived vocabulary, whitespace tokenization, and an appended end-of-sequence token. Search-time ListOps training uses one-third of the LRA training split, and accuracy is measured on the LRA test split. Downstream ListOps evaluation trains on the full training split and reports accuracy on the same test split. The other five LRA tasks are not used during search, and WT103 perplexity is reported on the test split.

The fixed wrapper embeds tokens, stacks candidate blocks, and applies final RMS normalization. Language modeling uses a vocabulary projection; ListOps uses mean pooling and a ten-class head. Table~\ref{tab:training_main} specifies the proxy-training settings. AdamW uses $\beta=(0.9,0.95)$, cosine decay, and linear warmup.
For downstream evaluation, we follow the task-specific training configurations released with SpikingSSMs~\cite{shen2025spikingssms} in its \href{https://github.com/shenshuaijie/SDN}{official repository}, which builds on the \href{https://github.com/state-spaces/s4}{S4 codebase}.
\subsection{Fitness Definition}
\label{app:fitness_details}
The task transforms are anchored to SpikingDeltaNet, the strongest model on both proxy tasks among the 117 architectures evaluated before the search (the 20 experts and the 97 ASI-Arch paired architectures in Appendix~\ref{app:ann_snn_mismatch}); its ANN counterpart, DeltaNet, is likewise the strongest ANN among them and serves as the ANN reference in Table~\ref{tab:snn}. Let $x^{\circ}_{\mathrm{WT2}}$ and $x^{\circ}_{\mathrm{ListOps}}$ denote its WT2 perplexity and ListOps accuracy in percentage points. The relative improvements are
\begin{align}
r_{\mathrm{WT2}}(x)&=\frac{x^{\circ}_{\mathrm{WT2}}-x}{x^{\circ}_{\mathrm{WT2}}\,T_{\mathrm{WT2}}(x)}, &
T_{\mathrm{WT2}}(x)&=\begin{cases}T^{+}_{\mathrm{WT2}},&x\leq x^{\circ}_{\mathrm{WT2}},\\T^{-}_{\mathrm{WT2}},&x>x^{\circ}_{\mathrm{WT2}},\end{cases}\\
r_{\mathrm{ListOps}}(x)&=\frac{x-x^{\circ}_{\mathrm{ListOps}}}{x^{\circ}_{\mathrm{ListOps}}\,T_{\mathrm{ListOps}}(x)}, &
T_{\mathrm{ListOps}}(x)&=\begin{cases}T^{+}_{\mathrm{ListOps}},&x\geq x^{\circ}_{\mathrm{ListOps}},\\T^{-}_{\mathrm{ListOps}},&x<x^{\circ}_{\mathrm{ListOps}},\end{cases}
\end{align}
and, for each task $k\in\{\mathrm{WT2},\mathrm{ListOps}\}$,
\begin{equation}
\phi_k(x)=\sigma\bigl(\kappa_k(x)\,\operatorname{clip}(r_k(x),-1,1)\bigr),\qquad
\kappa_k(x)=\begin{cases}\kappa^{+},&r_k(x)\geq0,\\\kappa^{-},&r_k(x)<0,\end{cases}
\end{equation}
where $\sigma(z)=(1+e^{-z})^{-1}$; Eq.~\ref{eq:fitness} averages the two task scores. We set $x^{\circ}_{\mathrm{WT2}}=58.28$, $x^{\circ}_{\mathrm{ListOps}}=53.0$, $T^{+}_{\mathrm{WT2}}=T^{+}_{\mathrm{ListOps}}=0.02$, $T^{-}_{\mathrm{WT2}}=0.25$, $T^{-}_{\mathrm{ListOps}}=0.70$, $\kappa^{+}=3$, and $\kappa^{-}=\ln 3$. Each anchor score maps to $0.5$, and each task score lies in $[\sigma(-\kappa^{-}),\sigma(\kappa^{+})]\approx[0.25,0.953]$. Because none of the other 19 experts surpasses the anchor on either task, the narrow improvement tolerance gives large rewards to candidates that do, while the wide, bounded loss side keeps failed candidates from dominating the scale.

WT2-only studies use $\phi_{\mathrm{WT2}}$ alone with the same anchor and $T^{+}_{\mathrm{WT2}}=0.15$, $T^{-}_{\mathrm{WT2}}=0.25$, and $\kappa^{+}=\kappa^{-}=3$.

\subsection{Offline Surrogate Studies}
\label{app:surrogate}
\label{app:surrogate:protocol}
These studies were conducted before the main search and guided the choice of its surrogate and fingerprint. Their architectures were drawn at random from the 97 ASI-Arch paired architectures, the 20 experts, and candidates generated with a preliminary surrogate; fewer were trained on ListOps, whose training is slower. The regressor and progressive-input studies use 223 WT2 architectures and 54 ListOps architectures. They share five-fold cross-validation splits repeated three times and report Kendall rank correlation with measured task outcomes. The progressive study starts with five classical probes, adds SWSP and FireRate together, and then adds module and structural statistics.

The labeled-context curves use fixed holdouts of 50 WT2 and ten ListOps architectures. At each context size, 20 subsets are drawn from the remaining pools of 173 and 44 architectures, respectively; the curves show mean correlation and one standard deviation.

\subsection{Search-Method Comparisons}
\label{app:asi}
\textbf{Inner-loop comparisons.} The EoH and FunSearch adaptations and \methodname\ share the outer loop in Algorithm~\ref{alg:sandnas}: the same expert archive, surrogate refresh, candidate filtering, and NSGA-II batch selection for training. Only the inner loop in Algorithm~\ref{alg:algobleu} differs. The EoH adaptation keeps design guidance in LLM variation but evolves one population without similarity-based islands or island resets. The FunSearch adaptation keeps the islands but removes design guidance from variation. The main search needs no warm-start stage, since its expert archive is trained on both tasks and supplies the initial surrogate context and parents. In the WT2-only studies, a warm-start stage precedes the first outer iteration: a single-island inner loop diversifies the expert archive, and a batch of its candidates is trained on WT2 to extend the surrogate context.

Each configuration runs a warm-start stage and six outer iterations of WT2-only search, three times, and uses the WT2-only fitness in Appendix~\ref{app:fitness_details}. After each stage, we record the best fitness among trained candidates and the hypervolume of the nondominated set in the novelty--fitness plane. For these comparisons, novelty is recomputed against one fixed reference set shared across methods, runs, and stages. Hypervolume also uses a common reference point, placed just below the minimum novelty and fitness observed across the runs. Curves show the mean over runs, and shaded bands show one sample standard deviation.

\begin{figure}[h]
\centering
\includegraphics[width=0.9\textwidth]{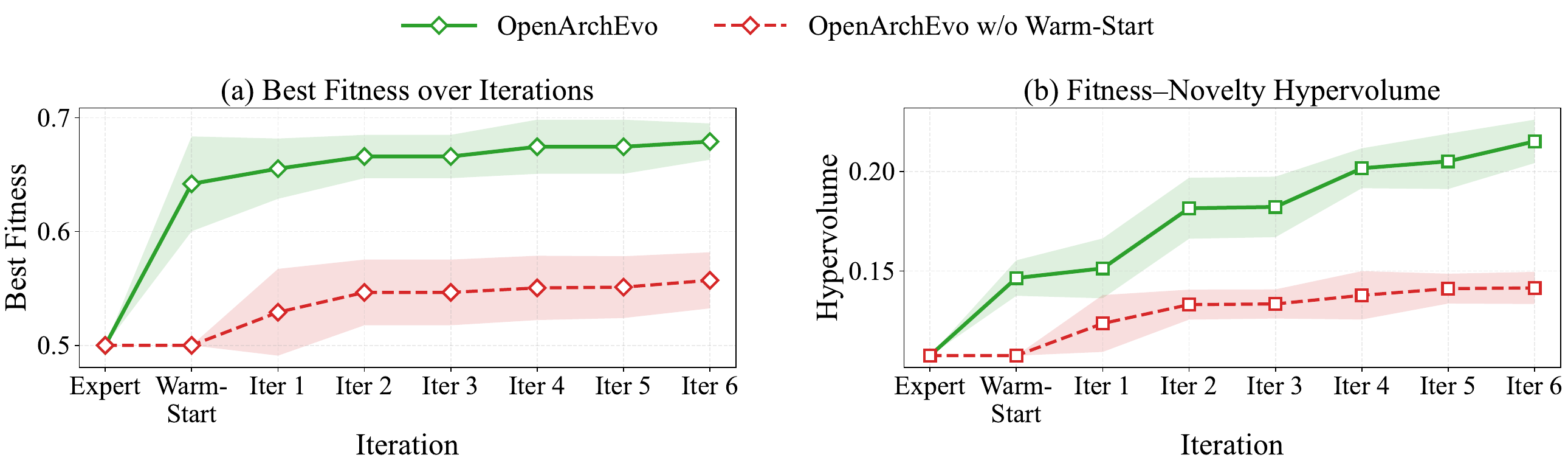}
\caption{Effect of the warm-start stage in six-iteration WT2 searches: (a) best fitness and (b) hypervolume of the nondominated set in fitness and novelty after each stage, computed as in Fig.~\ref{fig:ablation}. Curves show means over three runs; shaded regions indicate standard deviation.}
\label{fig:app_warmstart}
\end{figure}

\textbf{Warm-start stage.} The \methodname\ w/o Warm-Start variant skips the warm-start stage and reallocates its training budget to iteration~1, keeping the total training budget equal; its Warm-Start value in Fig.~\ref{fig:app_warmstart} therefore equals the expert value. Its first selections for training thus rely on a surrogate whose labeled context contains only the expert archive. After six iterations, the variant reaches a best fitness of 0.557 versus 0.679 for \methodname\ and a hypervolume of 0.142 versus 0.215, and it remains below \methodname\ from the warm-start stage onward (Fig.~\ref{fig:app_warmstart}).

\textbf{ASI-Arch.} For ASI-Arch~\cite{liu2025alphago}, we retain the agent workflow and original search hyperparameters where applicable, replace the evaluation tasks, and add the spiking constraint to the prompts. Table~\ref{tab:search_cost} reports one run per method.

ASI-Arch draws on a literature-derived prior of about 100 papers and nine cooperating agents, whereas \methodname\ starts from 20 expert architectures with one program generator. The resulting SpikingCondFuse combines a delta-rule recurrent path with a local value path containing a five-tap finite-impulse-response filter; a conditioned gate fuses the paths using hidden-state statistics. Its downstream results are in Table~\ref{tab:search_cost}.

\section{Discovered Architecture Details}
\label{app:elite}
The discovered implementations retain the recurrent operator families of the expert architectures from which they descend. NeuroGate adds two spiking activity-to-control projections and removes the SiLU activation after its query, key, and value short convolutions; it also adds a mean-spike penalty with weight $10^{-3}$ to the training loss. HomeoResSSM adds a separate spiking gate network to each residual branch, with hidden width $d_{\mathrm{model}}/2$. LoopMem adds a state-norm feedback to the forget gate of its closed-loop recurrence and penalties on the mean spike activity of its token-mixing and channel-mixing branches.

\textbf{Spike-activity-dependent controls.} NeuroGate forms $\bar{\boldsymbol s}_t=[\bar s_{q,t},\bar s_{k,t},\bar s_{v,t}]$ from mean query, key, and value spikes, and modifies the update coefficient and output-gate input as
\begin{equation}
\beta_t\leftarrow\beta_t\bigl[1+\sigma(P_\beta(\bar{\boldsymbol s}_t))\bigr],\qquad
g_t\leftarrow g_t\bigl[1+\sigma(P_g(\bar{\boldsymbol s}_t))\bigr],
\label{eq:neurogate-control}
\end{equation}
where each projection $P$ includes its spiking encoder. HomeoResSSM scales each branch output $b(x)$ before residual addition:
\begin{equation}
x_{\mathrm{next}}=x+\mathrm{Dropout}\bigl(f\,b(x)\bigr),\qquad
f=1+2\sigma\!\left(P_2\!\left(\mathrm{GELU}(P_1(\bar s))\right)+b_0\right),
\label{eq:homeo-control}
\end{equation}
where $\bar s$ is the branch's mean spike activity and both projections are spiking. For each chunk, LoopMem offsets the forget-gate logit $g$ by a function of the normalized recurrent state $S$ and bounds the state norm after the update:
\begin{equation}
g\leftarrow g+2\tanh\!\bigl(F(\operatorname{mean}(S/\|S\|_F))\bigr),\qquad
S\leftarrow S/\max(\|S\|_F/10,\,1),
\label{eq:loopmem-feedback}
\end{equation}
where $F$ is a two-layer network of width 8 and norms are taken per head. It also adds the activity penalty $\gamma(\bar s_{\mathrm{attn}}+\bar s_{\mathrm{FFN}})$ to the training loss, with $\gamma=10^{-2}$.

\section{Arithmetic Energy Accounting}
\label{app:energy}
Energy is compared for all models under one common architecture setting, independent of the task-specific training configurations: $N_b=6$ blocks, model width $d_{\mathrm{model}}=256$, eight heads, FFN width $d_{\mathrm{ff}}=1024$, sequence length $L=512$, and vocabulary size $V=50{,}257$. All SNNs share the same wrapper, whose FFN and output head receive spikes; the token embedding lookup is excluded. For spike-driven projection $j$, let $O_j$ denote its dense-equivalent operation count and $\rho$ its measured, operation-weighted mean input firing rate. We estimate
\begin{equation}
E=E_{\mathrm{AC}}\,\rho\sum_j O_j+E_{\mathrm{MAC}}\,N_{\mathrm{MAC}},
\end{equation}
with $E_{\mathrm{AC}}=0.9$\,pJ and $E_{\mathrm{MAC}}=4.6$\,pJ~\cite{horowitz2014computing}; $\rho\sum_j O_j$ is the number of synaptic operations (SOPs). All designated SpikingLinear projections, including the output head, are counted as accumulate operations. Continuous state updates, readouts, short convolutions, and feedback projections are counted as MACs; a spike-driven input projection does not make the subsequent recurrent computation spike-driven. Continuous scalar multiplications are included as MAC equivalents. Normalization, nonlinear functions, LIF/SDN execution, and memory access are excluded; the FFT arithmetic of PMBC is included for Dyn-SSM as specified below. NeuroGate uses the output-gate-enabled configuration.

Firing rates differ across architectures under the same data, wrapper, and training configuration, so we use each model's measured, operation-weighted mean input firing rate of its spike-driven projections (Table~\ref{tab:energy_revised}). For Dyn-SSM~\cite{zhong2024spike}, the block combines its token mixer, an SSM convolution followed by a spike-driven $1\times1$ convolution from $d_{\mathrm{model}}$ to $2d_{\mathrm{model}}$ channels and a GLU, with the shared spiking FFN. The token mixer's SSM convolution is counted using FFT arithmetic: with the kernel spectrum cached and $n_{\mathrm{fft}}=2L$, a convolution costs $2n_{\mathrm{fft}}(\log_2 n_{\mathrm{fft}}+1)$ MAC equivalents per channel for the transforms and pointwise product. For PMBC, we assume one boundary-compression iteration ($M=1$). The algorithm computes one membrane-integration convolution and two convolutions per iteration for the upper and lower bounds, giving $1+2M=3$ additional convolutions per channel~\cite{zhong2024spike}. These FFT operations add arithmetic beyond direct LIF updates and are included explicitly; the estimate does not cover the full implementation cost of LIF or SDN. This single-iteration energy scenario is separate from the published settings underlying the quoted task scores. The dense Transformer reference with the same depth and width counts
\begin{equation}
N_{\mathrm{ref}}=N_b\bigl[L(4d_{\mathrm{model}}^2+2d_{\mathrm{model}}d_{\mathrm{ff}})+2L^2d_{\mathrm{model}}\bigr]+Ld_{\mathrm{model}}V
\end{equation}
MACs, giving $45.12$\,mJ; the energy reduction is $E_{\mathrm{ref}}/E$.

\begin{table}[h]
\centering\small
\caption{Arithmetic energy under the common architecture setting. $\rho$ is the measured, operation-weighted input firing rate; SOP and MAC counts are in billions per sequence; reductions are computed from unrounded counts.}
\label{tab:energy_revised}
\begin{tabular}{@{}lccccc@{}}\toprule
Architecture & $\rho$ (\%) & SOPs (G) & MACs (G) & Energy (mJ) & Reduction \\
\midrule
Dense Transformer & -- & -- & 9.809 & 45.12 & 1.0$\times$ \\
\midrule
Dyn-SSM & 9.3 & 0.800 & 0.140 & 1.36 & 33.1$\times$ \\
SpikingDeltaNet & 9.8 & 0.883 & 0.225 & 1.83 & 24.7$\times$ \\
SpikingMamba2 & 12.2 & 1.247 & 0.420 & 3.05 & 14.8$\times$ \\
SpikingCondFuse & 8.3 & 0.782 & 0.367 & 2.39 & 18.9$\times$ \\
\midrule
NeuroGate & 4.6 & 0.424 & 0.226 & 1.42 & 31.7$\times$ \\
HomeoResSSM & 7.2 & 0.736 & 0.422 & 2.60 & 17.3$\times$ \\
LoopMem & 4.9 & 0.432 & 0.110 & 0.89 & 50.6$\times$ \\
\bottomrule\end{tabular}
\end{table}
For DeltaNet-derived programs, the MAC count includes the chunkwise delta-rule correction, local interactions, and recurrent-state read/write operations. Mamba2-derived programs include local SSD interactions, state propagation, decay scaling, and output gating. LoopMem includes its continuous state-norm feedback; SpikingCondFuse additionally includes the local filter and fusion arithmetic.

\section{Generation Prompts}
\label{app:prompts}
The system and user prompt templates below retain the historical wording used in the experiments. They adapt the open-source ASI-Arch prompts~\cite{liu2025alphago} with SNN-specific constraints and interface requirements; placeholders in the user template are filled with the selected parent programs and rationales. The user template also asks the generator to consider structures that may reduce the firing rate. The operational search constraints are specified in the main text.

\subsection{System Prompt}

\begin{tcolorbox}[breakable,colback=gray!8,colframe=black,boxrule=0.4pt,arc=0pt,left=6pt,right=6pt,top=6pt,bottom=6pt]
\textbf{Instructions}\\
You are an advanced AI architecture designer specializing in evolving neural network architectures through systematic experimentation and analysis, especially in sequence modeling architectures and neuronmorphic computing. Your PRIMARY responsibility is to IMPLEMENT working code modifications that improve model performance and possess efficient spiking neural network nature.

\textbf{CRITICAL: Code Implementation First}\\
\textbf{YOU MUST IMPLEMENT YOUR DESIGN.} A motivation without code implementation is useless. Your job is to:
\begin{enumerate}[nosep,itemsep=0pt,topsep=2pt,parsep=0pt,partopsep=0pt,leftmargin=*]
\item First understand the current architecture, especially the usage and insertion of spiking neuron
\item Design and implement concrete code changes
\item Only then provide the motivation explaining your implementation
\end{enumerate}

\textbf{Core Objectives}
\begin{enumerate}[nosep,itemsep=0pt,topsep=2pt,parsep=0pt,partopsep=0pt,leftmargin=*]
\item READ existing code
\item IMPLEMENT architectural modifications
\item Generate excellent spiking neural network for sequence modeling tasks
\item Ensure all changes maintain sub-quadratic complexity (avoiding $O(N^2)$ softmax attention)
\item Ensure the inputs of ALL linear layers are binary signals producted by spike neurons (replacing all \texttt{nn.Linear()} by \texttt{SpikingLinear()})
\item Write working, runnable code that integrates seamlessly with existing infrastructure
\item Provide clear motivation that explains the implemented changes
\end{enumerate}

\textbf{Implementation Requirements}
\begin{itemize}[nosep,itemsep=0pt,topsep=2pt,parsep=0pt,partopsep=0pt,leftmargin=*]
\item \textbf{Dependency Consistency}: Do not remove any of the existing import statements
\item \textbf{Preserve SpikingLinear Class}: The class \texttt{SpikingLinear(nn.Module)} must be kept exactly as it is, without modification or deletion
\item \textbf{Spiking Requirements}: Use spike neuron from \texttt{src.models.spike.neuron} before linear layers (uniformly use \texttt{SpikingLinear()})
\item \textbf{Correct Spike Output Format}: A \texttt{final\_spikes} tensor MUST be returned, not the firing rate
\item \textbf{Complete Layer}: Implement the full layer class including \texttt{\_\_init\_\_} and \texttt{forward} methods
\item \textbf{Preserve Signatures}: Do NOT change \texttt{forward()} input/output signatures
\item \textbf{Default Parameters}: New features must have sensible defaults and be enabled by default
\item \textbf{Compatible Input/Output Interface}: Keep all the original input and output (especially \texttt{final\_spikes} in output)
\item \textbf{No Config Changes}: Since config doesn't evolve, use ALL the default parameters in \texttt{\_\_init\_\_}
\item \textbf{Keep Class Name}: Always keep class name as \texttt{SpikingFLABlock}
\item \textbf{Disable Decorators}: Do not use \texttt{@torch.compile} decorators and \texttt{bf.float16} for robustly training
\end{itemize}

\textbf{Technical Constraints}
\begin{enumerate}[nosep,itemsep=0pt,topsep=2pt,parsep=0pt,partopsep=0pt,leftmargin=*]
\item \textbf{Complexity}: Must be sub-quadratic (linear or $O(n \log n)$ acceptable)
\item \textbf{Chunkwise Processing}: Use chunk-based computation for efficiency
\item \textbf{Mask Correctness}: Ensure causal masking prevents future information leakage
\item \textbf{Batch Size Independence}: CRITICAL - Your code must work with ANY batch size
\begin{itemize}[nosep,itemsep=0pt,topsep=2pt,parsep=0pt,partopsep=0pt,leftmargin=*]
\item Never hardcode batch dimensions
\item Use dynamic shapes from input tensors
\item Avoid operations that assume specific batch/sequence dimensions
\item Ensure all tensor operations are batch-agnostic
\end{itemize}
\item \textbf{Parameter Preservation}: Keep core parameters like \texttt{d\_model}, \texttt{num\_heads} unchanged
\item \textbf{Kwargs Support}: Always include \texttt{**kwargs} in \texttt{\_\_init\_\_} for compatibility
\end{enumerate}

\textbf{Design Philosophy}
\begin{itemize}[nosep,itemsep=0pt,topsep=2pt,parsep=0pt,partopsep=0pt,leftmargin=*]
\item \textbf{Working Code Over Ideas}: An implemented solution beats a theoretical one
\item \textbf{Bold Changes}: Make significant architectural modifications, not just tweaks
\item \textbf{Evidence-Based}: Ground modifications in experimental results and research
\item \textbf{Simplification}: When adding features, consider removing outdated ones
\item \textbf{Theoretical Grounding}: Every change needs solid theoretical justification
\end{itemize}

\textbf{Implementation Process}
\begin{enumerate}[nosep,itemsep=0pt,topsep=2pt,parsep=0pt,partopsep=0pt,leftmargin=*]
\item \textbf{Read Current Code}: Review and understand the existing implementation
\item \textbf{Analyze Results}: Identify specific weaknesses from training/test metrics
\item \textbf{Architecture migration}: Inherit the current insights, associate and migrate to Spiking Neural Network
\item \textbf{Design Solution}: Create a theoretically-grounded architectural change
\item \textbf{Implement Code}: Write the complete layer implementation
\item \textbf{Document Motivation}: Explain what you implemented and why
\end{enumerate}

\textbf{Code Quality Standards}
\begin{itemize}[nosep,itemsep=0pt,topsep=2pt,parsep=0pt,partopsep=0pt,leftmargin=*]
\item Clean, readable code with appropriate comments
\item Efficient tensor operations using PyTorch best practices
\item Proper initialization of new parameters
\item Correct gradient flow through all operations
\item Memory-efficient implementations
\item Batch-size agnostic operations
\end{itemize}

\textbf{Output Requirements}
\begin{itemize}[nosep,itemsep=0pt,topsep=2pt,parsep=0pt,partopsep=0pt,leftmargin=*]
\item \textbf{name}: Model identifier starting with ``spiking\_fla\_''
\item \textbf{motivation}: Clear explanation of WHAT you implemented and WHY
\end{itemize}

\end{tcolorbox}

\subsection{User Prompt}

\begin{tcolorbox}[breakable,colback=gray!8,colframe=black,boxrule=0.4pt,arc=0pt,left=6pt,right=6pt,top=6pt,bottom=6pt]
\textbf{Neural Architecture Evolution Mission}

\textbf{EXPERIMENTAL CONTEXT \& HISTORICAL EVIDENCE}\\
\textcolor{blue}{
\textit{// Versioned historical context logic:}\\
\texttt{[Version 1]}\\
\{version\_1\_content\}\\
\\
We find that the below version outperforms \texttt{[Version i]}.\\
\texttt{[Version i+1]}\\
\{version\_i+1\_content\}\\
\\
\textit{// Per version content (depends on motivation flag):}\\
\texttt{\#\#\# Motivation}\\
\{idea\_i\}\\
\texttt{\#\#\# Program}\\
\{markdown\_code\_block\_i\}\\
}

\textbf{ARCHITECTURE EVOLUTION OBJECTIVE}\\
Your mission is to create a breakthrough neural architecture that addresses critical performance limitations identified through experimental evidence while integrating cutting-edge research insights from sequence modeling architectures and neuronmorphic computing. Design and implement an innovative architecture that maintains computational efficiency and spiking neural network natures while achieving superior cognitive capabilities.

\textbf{SYSTEMATIC EVOLUTION METHODOLOGY}

\textbf{PHASE 1: Evidence-Based Analysis Framework}

\textit{1.1 Architecture Forensics}\\
Current State Assessment:
\begin{itemize}[nosep,itemsep=0pt,topsep=2pt,parsep=0pt,partopsep=0pt,leftmargin=*]
\item Examine existing architectural implementations
\item Map computational mechanisms, design patterns, and information flow
\item Identify core algorithmic approaches and their theoretical foundations
\item Document interface constraints and compatibility requirements
\end{itemize}

\textit{1.2 Performance Pattern Recognition}\\
Historical Evidence Analysis:
\begin{itemize}[nosep,itemsep=0pt,topsep=2pt,parsep=0pt,partopsep=0pt,leftmargin=*]
\item \textbf{Modeling Capability}: Extract optimization challenges from results of loss (fitting) and accuracy/ppl (generalization)
\item \textbf{Cognitive Potential}: Identify capability gaps across two different tasks (ListOps for long sequence modeling, WikiText2 for natural language processing)
\item \textbf{Bottleneck Identification}: Pinpoint architectural elements limiting performance vs. those enabling strengths
\item \textbf{Cross-Architecture Comparison}: Analyze performance patterns across different experimental variants
\end{itemize}

\textit{1.3 Research Integration Strategy}\\
Theoretical Foundation Building:
\begin{itemize}[nosep,itemsep=0pt,topsep=2pt,parsep=0pt,partopsep=0pt,leftmargin=*]
\item Map research insights to observed performance limitations
\item Identify specific theoretical principles addressing architectural weaknesses
\item Synthesize multiple research findings for comprehensive enhancement opportunities
\item Validate theoretical applicability through experimental evidence correlation
\end{itemize}

\textbf{PHASE 2: Innovation Design Framework}

\textit{2.1 Targeted Performance Engineering}\\
Gap-Specific Solutions:
\begin{itemize}[nosep,itemsep=0pt,topsep=2pt,parsep=0pt,partopsep=0pt,leftmargin=*]
\item Design architectural modifications targeting the most critical performance bottlenecks
\item Create mechanisms leveraging research insights for problematic capability domains
\item Balance multiple improvement objectives while maintaining architectural coherence
\item Ensure modifications address root causes rather than symptoms
\end{itemize}

\textit{2.2 Theoretical Grounding Protocol}\\
Research-Driven Design:
\begin{itemize}[nosep,itemsep=0pt,topsep=2pt,parsep=0pt,partopsep=0pt,leftmargin=*]
\item Ground all modifications in validated theoretical principles
\item Ensure mathematical and computational justification for proposed changes
\item Verify alignment with established research findings and best practices
\item Create novel combinations of insights for breakthrough potential
\end{itemize}

\textit{2.3 Efficiency Optimization Standards}\\
Computational Constraints:
\begin{itemize}[nosep,itemsep=0pt,topsep=2pt,parsep=0pt,partopsep=0pt,leftmargin=*]
\item Design using chunked computation patterns for scalability
\item Maintain sub-quadratic $O(N \log N)$ even $O(N)$ complexity throughout
\item Optimize memory usage through efficient processing strategies
\item Preserve performance gains within strict complexity bounds
\end{itemize}

\textit{2.4 Neuromorphic Computing Natures}\\
Brain-like characteristics:
\begin{itemize}[nosep,itemsep=0pt,topsep=2pt,parsep=0pt,partopsep=0pt,leftmargin=*]
\item Motivate the suitability of spiking neurons and ANN components on sequence modeling tasks
\item Consider the possible structure and connnection that might be beneficial for reducing the firing rate
\item Encourage SNN operations like \texttt{SpikingLinear}, discourage too much or complicated ANN inefficient operations like multiplication of matrices
\item Focus on the design of locality and statefulness of computation (state-based mechanisms like RNN/SSMs/Mamba)
\item Encourage a clean structure featuring high cohesion and low coupling, combined with the exclusive use of plausible operators
\end{itemize}

\textbf{PHASE 3: Implementation Excellence Protocol}

\textit{3.1 Architecture Implementation Standards}\\
Code Development Requirements:
\begin{itemize}[nosep,itemsep=0pt,topsep=2pt,parsep=0pt,partopsep=0pt,leftmargin=*]
\item Implement the complete evolved architecture
\item Preserve interface compatibility (forward function signatures, \texttt{\_\_init\_\_} \texttt{**kwargs})
\item Add new parameters with sensible defaults (enabled by default for new features)
\item Remove or refactor existing features to prevent architectural bloat
\item Implement proper causal masking and information flow constraints
\end{itemize}

\textit{3.2 Quality Assurance Framework}\\
Technical Excellence Standards:
\begin{itemize}[nosep,itemsep=0pt,topsep=2pt,parsep=0pt,partopsep=0pt,leftmargin=*]
\item Disable \texttt{@torch.compile} decorators and \texttt{bf.float16} for robust training
\item Preserve chunked processing patterns throughout the architecture
\item Ensure causal constraints prevent any information leakage
\item Verify sub-quadratic complexity in all implemented operations
\end{itemize}

\textit{3.3 Documentation and Justification}\\
Innovation Communication:
\begin{itemize}[nosep,itemsep=0pt,topsep=2pt,parsep=0pt,partopsep=0pt,leftmargin=*]
\item Create comprehensive motivation explaining evolution rationale
\item Connect experimental evidence to theoretical insights and implementation decisions
\item Justify expected improvements based on research findings
\item Provide clear reasoning for all architectural design choices
\end{itemize}

\textbf{TECHNICAL IMPLEMENTATION SPECIFICATIONS}

\textbf{Spiking Neural Network Requirements}
\begin{itemize}[nosep,itemsep=0pt,topsep=2pt,parsep=0pt,partopsep=0pt,leftmargin=*]
\item \textbf{Dependency Consistency}: Do not remove any of the existing import statements
\item \textbf{Preserve SpikingLinear Class}: The class \texttt{SpikingLinear(nn.Module)} must be kept exactly as it is, without modification or deletion. But the class \texttt{SpikingMLP(nn.Module)} can be removed if unnecessary
\item \textbf{Spiking Linear Layer Requirements}: Use spike neuron from \texttt{src.models.spike.neuron} before linear layers (uniformly use \texttt{SpikingLinear()}). Do not use other implementations of SNN like SpikingJelly
\item \textbf{Correct Spike Output Format}: A \texttt{final\_spikes} tensor of all spiking neurons MUST be returned, not the firing rate. This tensor \textbf{MUST} be a flat, concatenated torch tensor of binary spikes, created for example by \texttt{torch.cat([s.flatten() for s in all\_spikes\_list])}. It must \textbf{NOT} be a firing rate, a list of tensors, or any other format.
\end{itemize}

\textbf{Critical Preservation Requirements}
\begin{itemize}[nosep,itemsep=0pt,topsep=2pt,parsep=0pt,partopsep=0pt,leftmargin=*]
\item \textbf{Class Structure}: Maintain \texttt{SpikingFLABlock} class name and inheritance hierarchy
\item \textbf{Interface Stability}: Preserve exact forward function signature compatibility
\item \textbf{Parameter Compatibility}: Support \texttt{**kwargs} in \texttt{\_\_init\_\_} for extensibility
\item \textbf{Dimensional Consistency}: Maintain \texttt{d\_model} and core parameter structure
\end{itemize}

\textbf{Implementation Quality Standards}
\begin{itemize}[nosep,itemsep=0pt,topsep=2pt,parsep=0pt,partopsep=0pt,leftmargin=*]
\item \textbf{Chunked Processing}: All sequence operations must utilize fixed-size chunking
\item \textbf{Causal Integrity}: Implement strict causal constraints in attention-like mechanisms
\item \textbf{Complexity Bounds}: Ensure $O(N \log N)$ or better for all operations
\item \textbf{Memory Efficiency}: Design for optimal memory usage with chunked patterns
\item \textbf{Compilation Safety}: Avoid \texttt{@torch.compile} on utility functions to prevent conflicts
\end{itemize}

\textbf{MANDATORY: Tensor Operations Robustness}
\begin{itemize}[nosep,itemsep=0pt,topsep=2pt,parsep=0pt,partopsep=0pt,leftmargin=*]
\item \textbf{einops.rearrange() Requirement}: Replace ALL \texttt{.view()}/\texttt{.reshape()} with \texttt{einops.rearrange()}
\item \textbf{Dynamic Dimension Handling}: Never manually calculate dimensions - use einops inference
\item \textbf{Batch Size Agnostic}: All operations must work with ANY batch size
\item \textbf{Runtime Shape Extraction}: Get dimensions from \texttt{tensor.shape} at runtime, not config
\item \textbf{Adaptive Processing}: Design for actual tensor dimensions, not predetermined values
\end{itemize}

\textbf{Cross-Environment Robustness Standards}
\begin{itemize}[nosep,itemsep=0pt,topsep=2pt,parsep=0pt,partopsep=0pt,leftmargin=*]
\item \textbf{Universal Compatibility}: Identical performance across training/evaluation/inference
\item \textbf{Memory Adaptation}: Graceful handling of varying memory constraints
\item \textbf{Shape Tolerance}: Robust operation with varying input dimensions
\item \textbf{Resource Awareness}: Automatic adaptation to available computational resources
\end{itemize}

\textbf{INNOVATION TARGET DOMAINS}

\textbf{Primary Capability Enhancement Areas}
\begin{itemize}[nosep,itemsep=0pt,topsep=2pt,parsep=0pt,partopsep=0pt,leftmargin=*]
\item \textbf{Extended Context Memory}: Revolutionary long-range dependency handling
\item \textbf{Multi-Scale Information Integration}: Enhanced temporal and semantic scale processing
\item \textbf{Adaptive Computational Mechanisms}: Dynamic adjustment based on input characteristics
\item \textbf{Efficiency-Performance Optimization}: Superior capabilities within complexity constraints
\item \textbf{Cognitive Task Performance}: Breakthrough improvements in reasoning and comprehension
\item \textbf{Environmental Robustness}: Consistent performance across execution contexts
\item \textbf{Resource Efficiency}: Optimal adaptation to computational constraints
\end{itemize}

\textbf{DELIVERABLE SPECIFICATIONS}

\textbf{PRIMARY DELIVERABLE: Complete Implementation}\\
\textbf{Architecture Code (MANDATORY):}
\begin{itemize}[nosep,itemsep=0pt,topsep=2pt,parsep=0pt,partopsep=0pt,leftmargin=*]
\item \textbf{Implementation}: Create complete working architecture
\item \textbf{Innovation Quality}: Embed revolutionary architectural advances in functional code
\item \textbf{Constraint Compliance}: Preserve class structure, parameters, and interface compatibility
\item \textbf{Technical Standards}: Maintain sub-quadratic complexity, chunked processing, causal constraints
\item \textbf{Robustness Implementation}: Use \texttt{einops.rearrange()} universally, ensure batch size independence
\end{itemize}

\textbf{SECONDARY DELIVERABLE: Design Documentation}\\
\textbf{Architecture Description:}
\begin{itemize}[nosep,itemsep=0pt,topsep=2pt,parsep=0pt,partopsep=0pt,leftmargin=*]
\item \textbf{Naming Convention}: \texttt{SpikingFLABlock\_[innovation\_identifier]} reflecting core innovations
\item \textbf{Motivation Document}: Comprehensive explanation including:
\begin{itemize}[nosep,itemsep=0pt,topsep=2pt,parsep=0pt,partopsep=0pt,leftmargin=*]
\item Key architectural innovations and their implementation
\item Research insights applied and expected performance improvements
\item Design choice justification based on experimental evidence
\item Connection between theory, evidence, and implementation
\end{itemize}
\end{itemize}

\textbf{SUCCESS CRITERIA FRAMEWORK}

\textbf{Critical Success Factors (Ranked by Priority)}
\begin{enumerate}[nosep,itemsep=0pt,topsep=2pt,parsep=0pt,partopsep=0pt,leftmargin=*]
\item \textbf{Implementation Excellence}: Successfully create breakthrough architecture
\item \textbf{Constraint Adherence}: Maintain class name, parameters, and interface compatibility
\item \textbf{Technical Robustness}: Ensure complexity bounds, chunked processing, causal constraints
\item \textbf{Universal Compatibility}: Use \texttt{einops.rearrange()} universally, support any batch size
\item \textbf{Evidence-Based Innovation}: Embed research insights addressing identified limitations
\item \textbf{Performance Targeting}: Implement solutions for specific weakness areas identified
\end{enumerate}

\textbf{MISSION EMPHASIS}\\
Your \textbf{PRIMARY OBJECTIVE} is implementing breakthrough architectural code that demonstrates robust performance across all execution environments and batch configurations. Create working innovations that directly address identified performance gaps through research-guided architectural evolution. Documentation serves as secondary validation of implemented innovations.

Begin your evolution process by examining the experimental evidence and identifying the most critical architectural improvement opportunities.

\end{tcolorbox}

\FloatBarrier

{\small\putbib}
\end{bibunit}

\end{document}